\documentclass[sigconf]{acmart}

\AtBeginDocument{%
  \providecommand\BibTeX{{%
    \normalfont B\kern-0.5em{\scshape i\kern-0.25em b}\kern-0.8em\TeX}}}

\copyrightyear{2021}
\acmYear{2021}
\setcopyright{acmcopyright}\acmConference[KDD '21]{Proceedings of the 27th ACM SIGKDD Conference on Knowledge Discovery and Data Mining}{August 14--18, 2021}{Virtual Event, Singapore}
\acmBooktitle{Proceedings of the 27th ACM SIGKDD Conference on Knowledge Discovery and Data Mining (KDD '21), August 14--18, 2021, Virtual Event, Singapore}
\acmPrice{15.00}
\acmDOI{10.1145/3447548.3467123}
\acmISBN{978-1-4503-8332-5/21/08}

\usepackage{amsmath,amssymb,amsfonts}
\usepackage{graphicx}
\usepackage{textcomp}
\usepackage{xcolor}
\usepackage{algorithmic}
\usepackage[ruled]{algorithm2e}
\usepackage{todonotes}
\usepackage{color}
\usepackage{nowidow}
\usepackage{multirow}
\definecolor{mydarkblue}{rgb}{0,0.08,0.45}
\usepackage{enumitem}

\newcommand{\bx}{\mathbf{x}}
\newcommand{\mD}{\mathcal{D}}
\newcommand{\mL}{\mathcal{L}}
\newcommand{\mG}{\mathcal{G}}
\newcommand{\mS}{\mathcal{S}}

\let\svthefootnote\thefootnote

\begin{document}
\fancyhead{}

\title{Task-wise Split Gradient Boosting Trees for\\Multi-center Diabetes Prediction}

\author{Mingcheng Chen$^{1*}$, Zhenghui Wang$^{1*}$, Zhiyun Zhao$^{2\#}$, Weinan Zhang$^{1\#}$}
\author{Xiawei Guo$^3$, Jian Shen$^1$, Yanru Qu$^1$, Jieli Lu$^2$, Min Xu$^2$, Yu Xu$^2$}
\author{Tiange Wang$^2$, Mian Li$^2$, Wei-Wei Tu$^3$, Yong Yu$^1$, Yufang Bi$^{2}$, Weiqing Wang$^2$, Guang Ning$^2$}
\affiliation{\small $^{1}$Shanghai Jiao Tong University, Shanghai, China~~~$^{2}$Ruijin Hospital, SJTU School of Medicine, Shanghai, China~~~$^{3}$4Paradigm Inc., Beijing, China}
\affiliation{\{mcchen,felixwzh,wnzhang\}@apex.sjtu.edu.cn;~~~zzybrad@hotmail.com;~~~byf10784@rjh.com.cn \normalfont}








\renewcommand{\shortauthors}{M. Chen, Z. Wang, Z. Zhao et al.}
\renewcommand{\shorttitle}{Task-wise Split Gradient Boosting Trees for Multi-center Diabetes Prediction}

\begin{abstract}
Diabetes prediction is an important data science application in the social healthcare domain. 
There exist two main challenges in the diabetes prediction task: \emph{data heterogeneity} since demographic and metabolic data are of different types, \emph{data insufficiency} since the number of diabetes cases in a single medical center is usually limited.
To tackle the above challenges, we employ gradient boosting decision trees (GBDT) to handle \emph{data heterogeneity} and introduce multi-task learning (MTL) to solve \emph{data insufficiency}.
To this end, Task-wise Split Gradient Boosting Trees (TSGB) is proposed for the multi-center diabetes prediction task.
Specifically, we firstly introduce \emph{task gain} to evaluate each task separately during tree construction, with a theoretical analysis of GBDT's learning objective.
Secondly, we reveal a problem when directly applying GBDT in MTL, i.e., the \emph{negative task gain} problem.
Finally, we propose a novel split method for GBDT in MTL based on the task gain statistics, named \emph{task-wise split}, as an alternative to standard \emph{feature-wise split} to overcome the mentioned negative task gain problem.
Extensive experiments on a large-scale real-world diabetes dataset and a commonly used benchmark dataset demonstrate TSGB achieves superior performance against several state-of-the-art methods.
Detailed case studies further support our analysis of negative task gain problems and provide insightful findings.
The proposed TSGB method has been deployed as an online diabetes risk assessment software for early diagnosis.
\end{abstract}




\maketitle

\section{Introduction}
\let\thefootnote\relax\footnote{$*$ Co-first authors; $\#$ corresponding authors. }\addtocounter{footnote}{-1}\let\thefootnote\svthefootnote
As one of the deadliest diseases with many complications in the world, diabetes\footnote{We use diabetes to refer to type 2 diabetes in this paper.} is becoming one major factor that influences people's health in modern society \cite{breault2002data}.
To help prevent diabetes, predicting diabetes in the early stage according to demographic and metabolic data becomes an important task in the healthcare domain \cite{koh2011data}.
In this paper, we study the problem of diabetes prediction, specifically, predicting whether one person will be diagnosed with diabetes within three years based on the collected data, which can be regarded as a binary classification problem.
However, there exist two main challenges in this diabetes prediction task, i.e., \textit{data heterogeneity} and \textit{data insufficiency}.

\textit{Data heterogeneity} means the features contained in the collected data are of different types (e.g., ``glucose in urine'' is numerical while ``marriage'' is categorical) and distributions (e.g., ``gender'' tends to be evenly distributed while ``occupation'' normally follows a long-tail distribution).
Although deep neural networks (DNNs) have shown promising performance in vision and language domains \cite{goodfellow2016deep}, it is much harder to train DNNs with mixture types of input \cite{qu2018product}, especially when the input distribution is unstable \cite{ioffe2015batch}.
In contrast to DNNs, decision trees \cite{breiman2017classification} are insensitive to data types and distributions, and thus it is appealing to deal with heterogeneous data using decision trees \cite{breiman2017classification, qin2009dtu}.
More importantly, tree-based methods can implicitly handle the problem of features missing, which is common in medical follow-up data submitted by users, as we will mention in Section~\ref{sec:application}.
Beyond that, decision trees are also easy to visualization and further interpreted \cite{goodman2016clinical,walker2017decision} as we mention in Appendix~\ref{sec:appendix_B}, which is another superior advantage over neural networks.
Therefore, in this paper, to handle the \textit{data heterogeneity} challenge in diabetes prediction, we construct our method based on the gradient boosting decision trees (GBDT) \cite{friedman2002stochastic}, which is one of the most popular and powerful tree algorithms.

\textit{Data insufficiency} is another core challenge in the healthcare domain. Since data collection is costly in medical centers, the volume of data used to train models is usually limited, making it challenging to achieve adequate performance.
The data distribution of different medical centers can vary greatly. Thus training a model over the multi-center dataset cannot lead to a satisfactory average or separate prediction result.
Based on this observation, decoupling multi-center diabetes prediction and separately considering prediction tasks for each center is necessary.
Unfortunately, due to the data insufficiency, it is still hard to train high-performance models for every single center.
Multi-task learning (MTL) \cite{zhang2017survey} aggregates knowledge from different tasks to train a high-performance model. Based on this consideration, we are able to treat predictions for a single center as separate tasks and build the model leveraging knowledge shared among them to improve prediction results.

However, it is non-trivial to directly apply MTL on GBDT since most of the existing MTL methods are either feature-based or parameter-based \cite{ji2009accelerated, zhou2011clustered} but GBDT does not perform feature extraction and is a non-parametric model.
One existing solution is Multi-Task Boosting (MT-B) \cite{chapelle2010multi}, which simultaneously trains task-specific boosted trees with samples from each task, and task-common boosted trees with samples from all the tasks.
The final prediction of one task is determined by combining the predictions of both task-specific boosted trees and task-common boosted trees.
Although MT-B is easy to train and deploy, one significant drawback of MT-B is its high computational complexity, since independent trees for each task need to be learned besides the global one.

To avoid introducing additional computational complexity, seeking a more elegant way to address both challenges and combing MTL with GBDT is meaningful.
To begin with, we analyze that directly training one GBDT for all the tasks
may have a negative impact on specific tasks after a certain splitting since the task-wise difference is ignored.
For better understanding, we decompose the \textit{gain} into a summation over \textit{task gains} for each task and adopt the \textit{task gain} to measure how good the split condition is for each task at a node.
We demonstrate that the \textit{gain} being overall positive does not necessarily guarantee the \textit{task gains} of all tasks being also positive, which means that the greedy node split strategy directly based on \textit{gain} might be harmful to some tasks.

\begin{figure}
	\centering
	\includegraphics[width=0.985\linewidth]{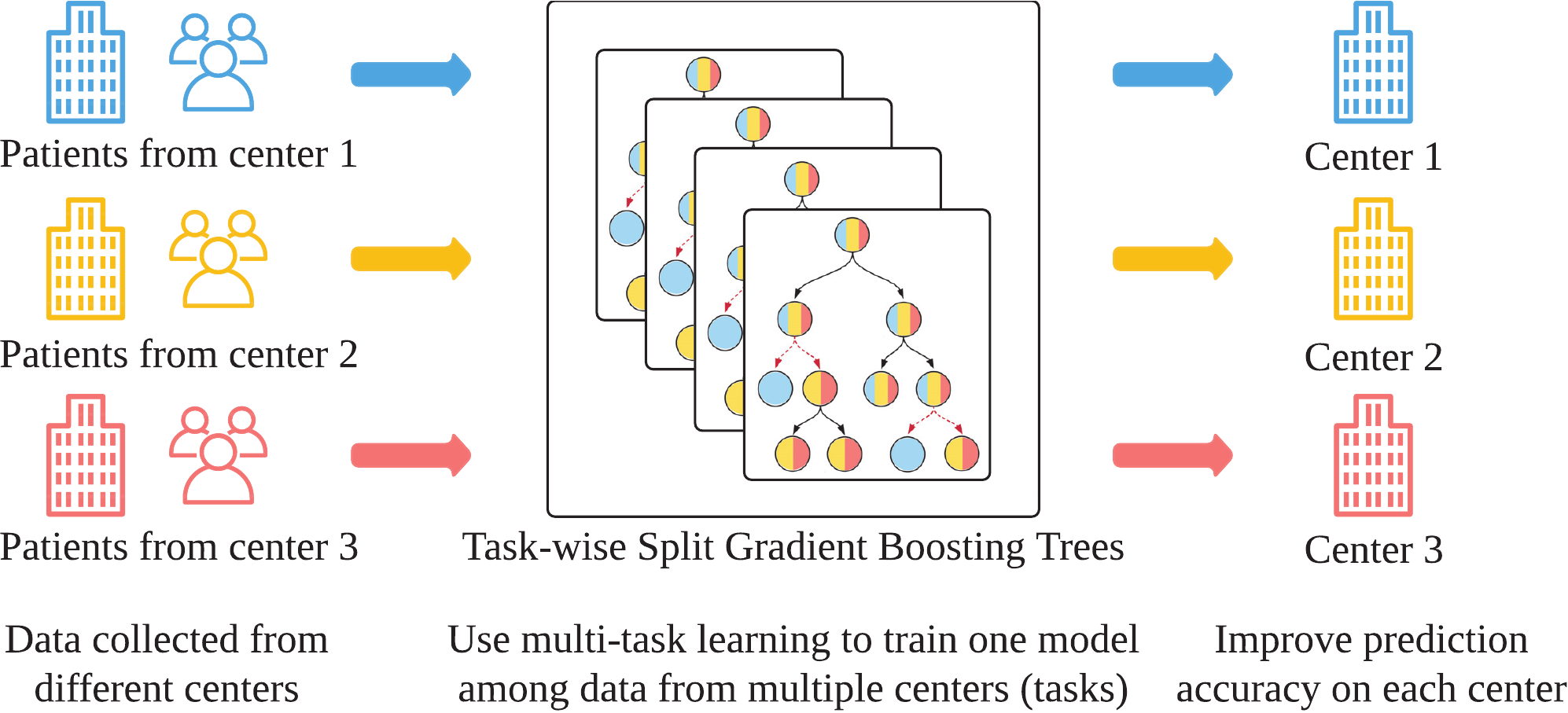}
	\vspace{-5pt}
	\caption{Workflow of TSGB for diabetes prediction.}
	\label{fig:Overview}
    \vspace{-15pt}
\end{figure}

To tackle this issue, inspired by MT-ET (Multi-Task ExtraTrees) \cite{simm2014tree}, we propose TSGB (Task-wise Split Gradient Boosting Trees).
TSGB introduces task-wise split according to task gain instead of traditional feature-wise split \cite{chen2016xgboost} to mitigate the \textit{negative task gain} problem while still keeping the same order of computational complexity as GBDT (all the tasks share trees).
Specifically, task-wise split separates tasks into two groups (see Fig.~\ref{fig:neg_task_gain/TSGB}), i.e., tasks with positive and negative gains.
In this way, some branches of the trees are only dedicated to a subset of tasks, which preserves the similarity between related tasks while alleviating the deficiency of sharing the knowledge between unrelated tasks.

The general workflow of applying our TSGB in diabetes prediction is illustrated in Fig.~\ref{fig:Overview}.
Experiments on multi-center diabetes prediction datasets and multi-domain sentiment classification dataset show the effectiveness of the proposed TSGB, compared with not only the tree-based MTL models \cite{chapelle2010multi,simm2014tree} but also several other state-of-the-art MTL algorithms \cite{ji2009accelerated,zhou2011clustered}.

The predictive model has been deployed as an online diabetes risk assessment software
to offer the patients key risk factors analysis and corresponding personalized health plan, helping early prevention and daily health management for healthy users.

To sum up, our contributions are mainly threefold:
\begin{itemize}[leftmargin=10pt]
    \item We analyze GBDT in the MTL scenario and introduce \textit{task gain} to measure how good the tree structure is for each task. To solve the \textit{negative task gain} problem, we propose a novel algorithm TSGB that effectively extends GBDT to multi-task settings. 
    \item We obtain 0.42\% to 3.20\% average AUC performance improvement on the 21 tasks in our diabetes prediction dataset, comparing with the state-of-the-art MTL algorithm. Our proposed TSGB is shown it can also be used in a wide range of non-medical MTL scenarios.
    \item We deploy TSGB on a professional assessment software, Rui-Ning Diabetes Risk Assessment, for fast and convenient diabetes risk prediction. The software already has around $48,000$ users from different organizations, such as physical examination centers, human resource departments, and insurance institutes.
\end{itemize}

\section{Preliminaries}\label{sec:naive-TSGB}
We provide a brief introduction to gradient boosting decision trees \cite{chen2016xgboost}.
Suppose we have a dataset $\mD=\{(\bx_i,y_i)\} (|\mD|=n,\bx_i\in\mathbb{R}^m,y_i\in\mathbb{R})$ of $n$ samples with $m$-dimensional features. The predicted label of GBDT given by the function $\phi$ is the sum of all the $K$ additive trees:
\begin{align}
\label{eq:add}
\small
\hat{y}_i=\phi(\bx_i)=\sum_{k=1}^{K}f_k(\bx_i),\  f_k\in\mathcal{F},
\end{align}
\begin{align}
\small
\text{where~}\mathcal{F}=\{f(\bx)=w_q(\bx)\}(q: \mathbb{R}^m\to L, w\in\mathbb{R}^{|L|})
\end{align}
is the space of regression trees (CART \cite{breiman2017classification}), $q$ is the tree structure which maps a sample $\bx$ to the corresponding leaf index in the tree with $|L|$ leaves, and $w$ is the leaf weight.
Each $f_k$ is an independent tree with its own structure $q_k$ and leaf weight $w_{q_k}$.
The $K$ functions (trees) will be learned by minimizing the \textit{regularized} objective \cite{chen2016xgboost}:
\begin{align}
\small
\mL(\phi)&
=\sum_{i=1}^n l(\hat{y}_i,y_i)+\sum_{k=1}^K\Omega(f_k),\label{eq:original-obj}
\end{align}
where $l(\hat{y},y)$ is the loss function (e.g., MSE, logloss), and $\Omega(f)$ is a regularization term that penalizes the complexity of $f$ to alleviate the over-fitting problem.
Specifically, $\Omega(f)$ penalizes the number of leaves as well as the weight values  \cite{johnson2014learning,chen2016xgboost}:
\begin{align}
\small
\Omega(f)=\gamma |L|+\frac{1}{2}\lambda\|w\|^2_2 \label{eq:regularization}.
\end{align}

Following the GBM \cite{friedman2001greedy} framework, the $K$ functions in Eq.~\eqref{eq:add} are learned additively to minimize the objective in Eq.~\eqref{eq:original-obj}.
With some transformation and simplification (see details in \cite{chen2016xgboost}), the $s$-th tree is learned by minimizing the following objective $\tilde{\mathcal{L}}^{(s)}$ as
\begin{align}
\small
\tilde{\mathcal{L}}^{(s)}=\sum_{i=1}^n \Big[g_i f_s(\mathbf{x}_i)+\frac{1}{2}h_i f_s^2(\mathbf{x}_i)\Big]+\Omega(f_s),\nonumber\\
g_i=\frac{\partial l(y_i,\hat{y_i}^{(s-1)})}{\partial \hat{y_i}^{(s-1)}},\ \  h_i=\frac{\partial^2l(y_i,\hat{y_i}^{(s-1)})}{\partial(\hat{y_i}^{(s-1)})^2}\nonumber,
\end{align}
where $g_i$ and $h_i$ are the first-order and second-order gradient on the loss function.
Note that each sample will be mapped to a leaf via $f_s$, thus we define $I_j=\{i|q_s(\bx_i)=j\}$ as the indices set of training samples at leaf $j$ where $q_s(\bx)$ is the corresponding tree structure.
Recall the definition of $\Omega(f_s)$ in Eq.~\eqref{eq:regularization}, we have:
\begin{align}
\small
\tilde{\mathcal{L}}^{(s)}&=\sum_{j=1}^L \Big[(\sum_{i\in I_j}g_i)w_j+\frac{1}{2}(\sum_{i\in I_j}h_i)w_j^2 \Big]+\gamma L+\frac{1}{2}\lambda \sum_{j=1}^L w_j^2\nonumber\\
&=\sum_{j=1}^L \Big[(\sum_{i\in I_j}g_i)w_j+\frac{1}{2}(\sum_{i\in I_j}h_i+\lambda)w_j^2 \Big]+\gamma L\label{eq:obj-s-origin}.
\end{align}
Denoting $G_j=\sum_{i\in I_j}g_i$ and $H_j=\sum_{i\in I_j}h_i$, the optimal $w^*_j$ for leaf $j$ is easy to calculate since $G_jw_j+\frac{1}{2}(H_j+\lambda)w_j^2$ is a single variable quadratic function for $w_j$, thus the optimal $w_j^*$ is
\begin{align}\label{eq:weight}
\small
w_j^*=-\frac{G_j}{H_j+\lambda}=-\frac{{\sum_{i\in I_{j}}g_i}}{{\sum_{i\in I_{j}}h_i}+\lambda}.
\end{align}



Although the optimal value of $\tilde{\mathcal{L}}^{(s)}$ given tree structure $q_s(\bx)$ can be calculated,
to make a trade-off between computational complexity and model performance, a greedy strategy that constructs a tree starting from a single leaf (root) and splitting the leaf into two child leaves iteratively is commonly used \cite{breiman2017classification,johnson2014learning,chen2016xgboost}.
The samples at a leaf will be separated by the split condition defined as a threshold value of one feature, which is the so-called feature-wise split.
Such a greedy search algorithm is included in most GBDT implementations \cite{ridgeway2007generalized,pedregosa2011scikit,chen2016xgboost}, it selects the best split node by node, and finally construct a decision tree.

Formally, to find the best split condition on an arbitrary leaf $p$, let $I$ be the sample set at leaf $p$, and $I_{L}$ and $I_{R}$ are the samples for left and right child leaves ($p_L$ and $p_R$) after a split.
The corresponding negative loss change after the split, denoted as \textit{gain} $\mathcal{G}$, is
\begin{align}
\small
\mG
=&-\sum_{i\in I_{L}}g_iw_{p_L}^*-\frac{1}{2}(\sum_{i\in I_{L}}h_i+\lambda)w_{p_L}^{*^2}-\sum_{i\in I_{R}}g_iw_{p_R}^*\nonumber\\
&-\frac{1}{2}(\sum_{i\in I_{R}}h_i+\lambda)w_{p_R}^{*^2}+\sum_{i\in I}g_iw_p^*+\frac{1}{2}(\sum_{i\in I}h_i+\lambda)w_p^{*^2}-\gamma\label{eq:gain_all}\\
=&\frac{1}{2} \Big[
\frac{({\sum_{i\in I_{L}}g_i})^2}{{\sum_{i\in I_{L}}h_i}+\lambda}
+\frac{({\sum_{i\in I_{R}}g_i})^2}{{\sum_{i\in I_{R}}h_i}+\lambda}
-\frac{({\sum_{i\in I}g_i})^2}{{\sum_{i\in I}h_i}+\lambda} \Big]-\gamma,\nonumber
\end{align}
where $w_{p}^{*},w_{p_L}^{*},w_{p_R}^{*}$ are optimal weights (defined in Eq.~\eqref{eq:weight}) for leaf $p,p_L,p_R$, respectively.
There is an optimal split found for each feature by enumerating all the possible candidate feature values and picking one with the highest gain.

\section{Negative Task Gain Problem} \label{sec:negtive_task_gain}
We find that the tree structure learned by GBDT can be harmful to a subset of tasks when the MTL technique is directly applied. When training vanilla GBDT on multi-task data, where samples from different tasks may be far from identically distributed (e.g., multi-center diabetes dataset), the objective is to improve its average performance over all the tasks against individual learning. To be specific, since the objective is defined on all the training instances, GBDT will pick features that are generally ``good'' for all the tasks in the feature-wise splitting process of growing a single tree.

To illustrate this finding, we need to analyze the tree structure measurement in task level. Assume there are $T$ learning tasks in the MTL scenario with the whole dataset $\mD$ divided into $T$ parts ($\mD=\mD_1\cup\mD_2\cup \ldots \cup\mD_T \text{ and } \mD_i\cap\mD_j=\varnothing, i\neq j$).
For each task $t$, denote the samples belonging to it as $\mathcal{S}^t$, thus have $\mD_t=\{(\bx_i,y_i)|i\in\mathcal{S}^t\}$.
We now introduce a new metric, \textit{task gain} ($\mG_t, t\in\{1,2,...,T\}$), to measure how good a feature-wise split is to each task.




Considering all the $T$ tasks explicitly at each leaf, the learning objective at $s$-step in Eqs.~\eqref{eq:original-obj} and \eqref{eq:obj-s-origin} can be rewritten as
\begin{align}
\small
&\tilde{\mathcal{L}}^{(s)}
=\sum_{j=1}^L \Big\{(\sum_{i\in I_j}g_i)w_j+\frac{1}{2}(\sum_{i\in I_j}h_i+\lambda)w_j^2\Big\}+\gamma L \nonumber\\
=&\sum_{j=1}^L\Big\{(\sum_{t=1}^{T} \sum_{i\in I_j\cap\mS^t}g_i)w_j+\frac{1}{2}(\sum_{t=1}^{T} \sum_{i\in I_j\cap\mS^t}h_i+\lambda)w_j^2\Big\} \nonumber + \gamma L\nonumber\\
=&\sum_{j=1}^L\Big{\{}\sum_{t=1}^{T}\Big[G_j^tw_j
+\frac{1}{2}(H_j^t+\frac{|I_j\cap\mS^t|}{|I_j|}\lambda)w_j^2\Big]\Big{\}}+\gamma L,
\end{align}
where $G_j^t = \sum_{i\in I_j\cap\mS^t}g_i , H_j^t=\sum_{i\in I_j\cap\mS^t}h_i$.
Then according to the objective above, we can decompose $\mG$ in Eq.~\eqref{eq:gain_all} by $\mG = \sum_{t=1}^{T}\mG_t$ as
\begin{align}
\small
\mG_t
=&-\sum_{i\in I_{p_L}^t}g_iw_{p_L}^*-\frac{1}{2} \Big(\sum_{i\in I_{p_L}^t}h_i+\frac{|I_{p_L}^t|}{|I_{p_L}|}\lambda \Big)w_{p_L}^{*^2}\nonumber\\
&-\sum_{i\in I_{p_R}^t}g_iw_{p_R}^*-\frac{1}{2} \Big(\sum_{i\in I_{p_R}^t}h_i+\frac{|I_{p_R}^t|}{|I_{p_R}|}\lambda \Big)w_{p_R}^{*^2}\nonumber\\
&+\sum_{i\in I_{p}^t}g_iw_{p}^*+\frac{1}{2} \Big(\sum_{i\in I_{p}^t}h_i+\frac{|I_{p}^t|}{|I_{p}|}\lambda \Big)w_{p}^{*^2}-\frac{|I_{p}^t|}{|I_{p}|}\gamma,\label{eq:gain_decomposition}
\end{align}
where $I^t_p=I_{p}\cap\mS^t$ denotes the set of samples from task $t$ in leaf $p$, as well as $I^t_{p_L}$ and $I^t_{p_R}$.

With the above decomposition of the original gain $\mG$ of a feature-wise split at a leaf in GBDT, we obtain the task gain $\mG_t$ for each task.
The task gain $\mG_t$ represents how good the specific feature-wise split at this leaf is for task $t$. The larger $\mG_t$ is, the better the split at this leaf is for task $t$.
When $\mG_t$ is negative, it means the feature-wise split at this leaf will actually increase part of the objective loss consisting of samples in task $t$ as this leaf: $\sum_{i\in I_p^t}l(\hat{y}_i,y_i)$, which is the opposite of the optimization objective.

\begin{figure}[tbp]
\centering
\includegraphics[width=0.5\linewidth]{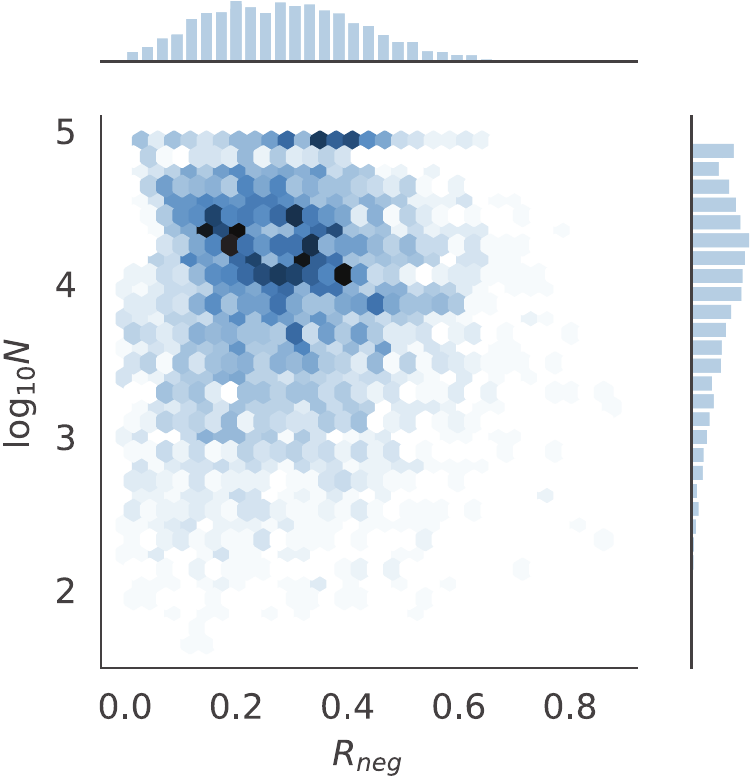}
\caption{
Distribution of non-leaf nodes' $R_{\text{neg}}$ with logarithm, when trained traditional GBDT on multi-center diabetes dataset. A spot in darker blue has more nodes. }
\label{fig:xgb_hist_neg_task_gain_ratio}
\vspace{-0.5cm}
\end{figure}

In GBDT, we search over all the feature-wise split conditions and select the one with the highest gain $\mG^*$ at a leaf.
Consider the decomposition in Eq.~\eqref{eq:gain_decomposition}, we can conclude that there is no guarantee that the optimal feature-wise split is a good split for all the tasks.
Formally, according to the greedy algorithm for split finding in GBDT, we have
\begin{align}
\small
\{\mG_1^*,\mG_2^*,...,\mG_T^*\}=\mathop{\arg\min}_{\{\mG_1,\mG_2,...,\mG_T\}}\  \sum_{t=1}^{T}\mG_t=\mathop{\arg\min}_{\{\mG_1,\mG_2,...,\mG_T\}}\   \mG, \nonumber
\end{align}
at a leaf, but unfortunately,
\begin{align}
\sum_{t=1}^{T}\mG_t^*>0\nRightarrow \mG_t^*>0, \forall t\in\{1,2,...T\}.\nonumber
\end{align}
We dub this observation \textit{negative task gain problem}.
For the tasks with the task gain $\mG_t^*<0$, although the feature-wise split is good in general ($\mG^*>0$), the newly constructed tree structure is even worse.
Empirically, we find there are about 96.47\% of nodes in GBDT that have negative task gains when trained on our diabetes dataset.

To get a better measurement for ``how good is a feature-wise split in multi-task settings'', we introduce \emph{negative task gain ratio} as
\begin{align}\label{eq:R_neg}
\small
R_{\text{neg}}=\sum_{t\in\{i|\mG_i<0\}}\frac{|I_{p}^t|}{|I_{p}|}
\end{align}
to indicate the severity of the negative task gain problem, where $|I_{p}^t|$ is the number of samples with negative task gains, $|I_p|$ is the total number of samples at node $p$.
We plot the distribution of $R_{\text{neg}}$ in Fig.~\ref{fig:xgb_hist_neg_task_gain_ratio} and find that (i) a large amount of nodes have $0.2 < R_{\text{neg}} < 0.4, 10^4 < N < 10^5$, which means the greedy search algorithm in GBDT is far from optimum in multi-task settings.
(ii) Nodes with more samples are more likely to have larger $R_{\text{neg}}$, which means in the early stage of training, nodes closer to the root are more likely to find a harmful feature-wise split. And different tasks sharing the same harmful tree structure will, of course, lead to performance decline.
(iii) There are 11.24\% nodes have $R_{\text{neg}}>0.5$, which means a minority of the tasks dominates the feature-wise split, and the other tasks will gain better results if the split is not performed.

\begin{figure}[tbp]
    \centering
    \begin{minipage}[t]{0.45\linewidth}
    \centering
    \includegraphics[width=1\linewidth]{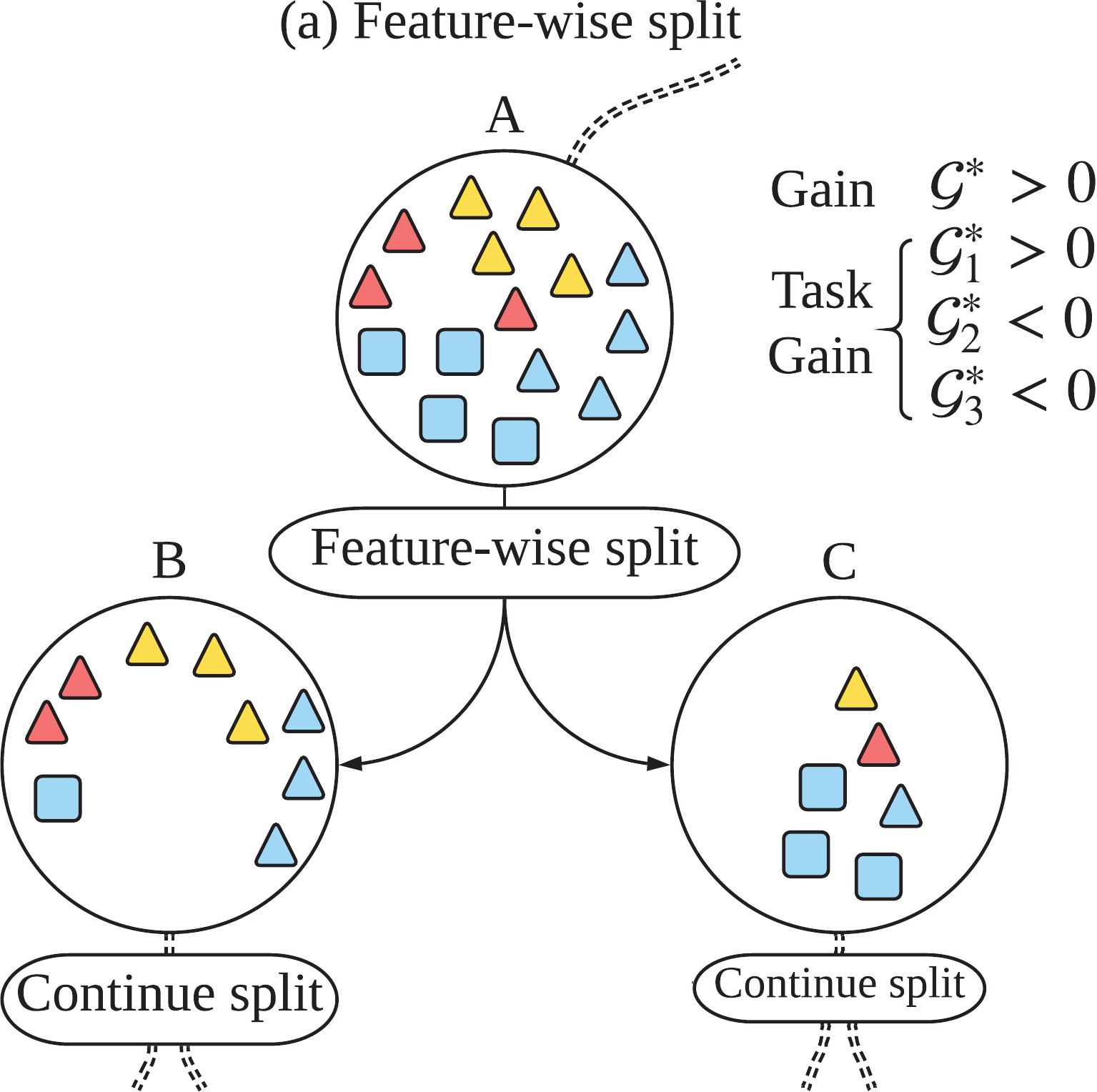}
    \end{minipage}
    \begin{minipage}[t]{0.45\linewidth}
    \centering
    \includegraphics[width=1\linewidth]{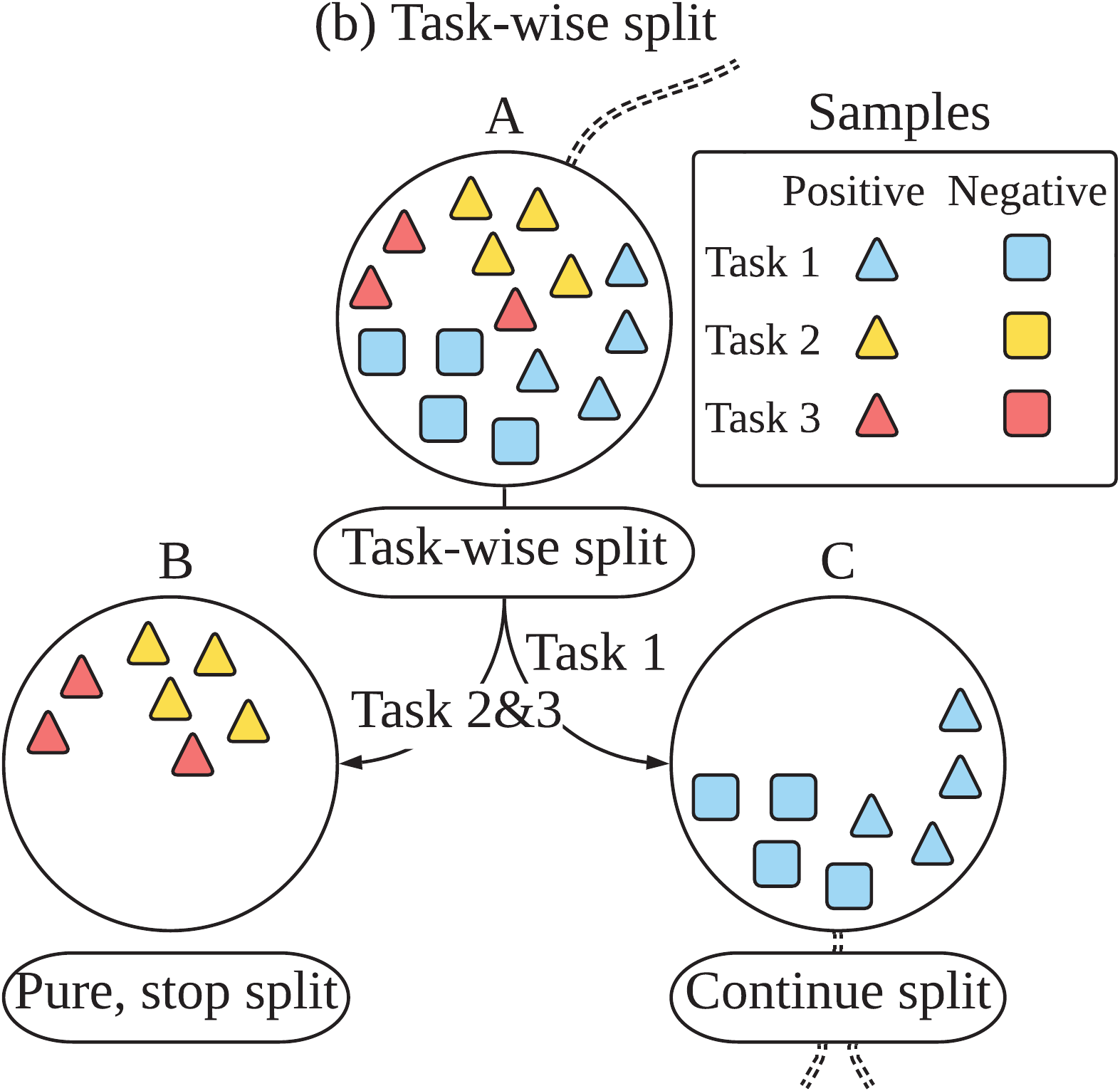}
    \end{minipage}
    \caption{(a) Illustration of negative task gain problem caused by a traditional feature-wise split, (b) while proposed TSGB can handle such problem with a task-wise split.}
    \label{fig:neg_task_gain/TSGB}
\end{figure}

To better illustrate this problem, we show a simple but common case found in GBDT in Fig.~\ref{fig:neg_task_gain/TSGB} (a).
At node $A$, samples of tasks $2,3$ are already pure (all positive), while the positive and negative samples of task $1$ are still mixed.
The optimal split condition found here successfully divides task $1$'s samples into two branches, and the right branch has the most negative samples of task $1$ while the left branch has most of the positive ones.
Unfortunately, some samples of tasks $2,3$ are also divided into the right branch, although they are positive samples.
In such a case, we find the optimal $\mG^*$ and $\mG_1^*>0$, but leave the rest of the tasks with negative gains ($\mG_2^*<0$ and $\mG_3^*<0$).



	
	

\begin{algorithm}[tbp]
    \small
	\caption{Task-wise Split Gradient Boosting Trees}\label{alg:TSGB}
	\LinesNumbered
	\KwIn{$\mD=\{(\bx_i,y_i)\}$, training data from $T$ tasks}
	\KwIn{$K$, number of boosted trees}
	\KwIn{$R$, maximum ratio of samples with negative task gain}
	initialize $f_1$\\
	
	\For{$k=2$ {\bfseries to} $K$ }{
		Calculate $\hat{y_i}^{(s)}$ by Eq.~\eqref{eq:add}\\
		\While{not meet the split stop criterion}{
		Find the best feature-wise split rule $r$ greedily at a leaf $p$\\
		Calculate Corresponding task gain $\mG_t, t\in\{1,2,...,T\}$ defined in Eq.~\eqref{eq:gain_decomposition}\\
		\If{$R_{\text{neg}}>R$}{Split samples of task $t\in\{i|\mG_i<0\}$ to left branch\\
		Split samples of task $t\in\{i|\mG_i\geq0\}$ to right branch\\
		 }\Else{Split samples following split rule $r$ }
		}
	}
	\KwOut{$\phi(\cdot)=\sum_{k=1}^{K}f_k(\cdot)$}
\end{algorithm}

\section{Task-wise Split Gradient Boosting Trees}
The ultimate objective of MTL is to improve the model's performance on all the tasks, while the aforementioned \textit{negative task gain} problem makes the traditional GBDT not suitable for MTL. To make full use of the data of all tasks through MTL and extend GBDT to multi-task settings, we propose Task-wise Split Gradient Boosting Trees (TSGB). The key idea of TSGB is that we avoid severe \textit{negative task gain} problem  by conducting a \textit{task-wise split} instead of \textit{feature-wise split} at nodes with high negative task gain ratio $R_{\text{neg}}$.

We follow the main procedure of GBDT. However, when the best feature-wise split condition is found at an arbitrary leaf $p$, the task gain $\mG_t$ for each task $t$ is calculated.
Since most of the nodes in GBDT has the negative task gain problem (as discussed in Fig.~\ref{fig:xgb_hist_neg_task_gain_ratio}).
We can handle this problem by introducing the task-wise split.

If the negative task gain ratio $R_{\text{neg}}$ of node $p$, as defined in Eq.~\eqref{eq:R_neg}, is higher than a threshold ratio $R$ (i.e., $R_{\text{neg}} > R$ meets),
instead of splitting the leaf feature-wisely using the found split condition, TSGB performs a \textit{task-wise split} of samples, splits the samples of tasks with negative task gain to the left branch and those with positive task gain to the right branch.
Alg.~\ref{alg:TSGB} is a pseudo-code for TSGB, Fig.~\ref{fig:neg_task_gain/TSGB} (right) provides an illustration of proposed task-wise split.
$R$ is considered as a hyperparameter in practice, which is set to different values for different MTL datasets.


A key characteristic of TSGB is that it is task-level objective-oriented while training all the tasks in the same trees in a homogeneous MTL setting, which makes TSGB easy to train and elegant in MTL. The empirical results also show the effectiveness of TSGB.
Previous works either ignore the task-specific objective by simply splitting the tasks with pre-defined task features on randomly selected leaf nodes \cite{simm2014tree} or train both task-common trees and task-specific trees at the same time. The former can not make full use of the correlation of different task data, while the latter always derive a huge redundant model since $T+1$ forests are needed \cite{chapelle2010multi}.

What if replacing task-wise split by selecting the sub-optimal feature-wise split condition at a node with lower $R_{\text{neg}}$ so that more of the tasks have positive task gain?
We argue that (i) the primary cause of the negative task gain problem comes from the difference of feature distributions on different tasks.
Moreover, this problem cannot be solved by traditional greedy search feature-wise split since its underlying assumption is identical data distribution.
There is an irreparable gap between the assumptions of GBDT feature split and MTL.
(ii) The computational complexity of task gain calculation under sub-optimal feature-wise split conditions is much higher.
Variables are not provided by the original GBDT. Thus if we want to calculate the task gain under sub-optimal feature-wise split, additional computation is needed.
This problem even becomes worse given that the searching of sub-optimal feature-wise split brings external complexity.

Accordingly, we only perform a task-wise split when calculated task gain under the optimal feature-wise split condition indicates that the negative task gain problem meets a certain condition.
\section{Experiments}
In this section, we empirically study the performance of the proposed TSGB.
We compare TSGB with several state-of-the-art tree-based and other MTL algorithms on our diabetes dataset and the multi-domain sentiment dataset. To get deeper insights into how task-wise split helps improve the performance, we also discuss a case study on a specific task.

\subsection{Dataset}
We first evaluate TSGB on a multi-center diabetes dataset provided by the China Cardiometabolic Disease and Cancer Cohort (4C) Study, which was approved by the Medical Ethics Committee of Ruijin Hospital, Shanghai Jiao Tong University. 
The dataset is collected from the general population recruited from 21 medical centers in different geographical regions of China, including the baseline data from 2011 to 2012 and the follow-up data from 2014 to 2016 of 170,240 participants. Each center contributes the data with the size ranging from 2,299 to 7,871.
At baseline and follow-up visits, standard questionnaires were used to collect demographic characteristics, lifestyle, dietary factors, and medical history. 
We finally obtained 100,000 samples from the 21 different medical centers for TSGB evaluation with data cleaning and pre-processing. Each of the samples retains the most important numerical and categorical features of 50 dimensions.

To further claim the effectiveness of TSGB under a non-medical scenario, we also conduct an empirical study on a commonly used MTL benchmark dataset, the multi-domain sentiment dataset\footnote{Sentiment dataset: \url{https://www.cs.jhu.edu/~mdredze/datasets/sentiment/}} \cite{blitzer2007biographies}. This dataset is a multi-domain sentiment classification dataset, containing positive and negative reviews from four different (product) domains from Amazon.com. The four product domains are books, DVDs, electronics, and kitchen appliances. Following \cite{chen2012marginalized}, we use the 5000 most frequent terms of unigrams and bigrams as the input.

\subsection{Baselines}
All the compared models are listed as follows. 
\begin{itemize}[leftmargin=10pt]
	\item \textbf{ST-GB} (Single Task GBDT) 
	 trains a GBDT model for each task separately.
	\item \textbf{GBDT} (Sec.~\ref{sec:naive-TSGB}) trains a  GBDT model on the whole dataset of all tasks.
	\item \textbf{MT-ET} (Multi-Task ExtraTrees) \cite{simm2014tree} is a tree-based ensemble multi-task learning method based on Extremely Randomized Trees.
	\item \textbf{MT-TNR} \cite{ji2009accelerated} is a linear MTL model with Trace Norm Regularization.
	\item \textbf{MT-B} (Multi-Task Boosting) \cite{chapelle2010multi} is an MTL algorithm with boosted trees. It trains task-common forest $F_0$ on all tasks, and trains task-specific boosted forest $F_t,i\in\{1,2,...,T\}$ on each task separately. The final output of sample $\bx$ of task $t$ is $F_0(\bx)+F_t(\bx)$.
	\item \textbf{CMTL} \cite{zhou2011clustered} is a clustered MTL method that assumes the tasks may exhibit a more sophisticated group structure.
	\item \textbf{TSGB$_{\lambda}$} is a variant of TSGB proposed by us. Instead of using a threshold $R$, it decides whether to conduct a task-wise split with a fixed probability $\lambda\in[0, 1]$. TSGB$_{\lambda}$ picks a node with a certain probability and sort the tasks, then split the samples task-wisely as how TSGB do.
	\item \textbf{TSGB} is the novel method proposed in this paper. It decides whether to perform a task-wise split by comparing a threshold ratio $R$ with the negative task gain ratio $R_{\text{neg}}$ of the current node. Then it separates samples task-wisely according to the positive and negative of their task gains.
\end{itemize}
For a fair comparison, all the boosting tree models used in our experiments are implemented based on XGBoost \cite{chen2016xgboost}, which is one of the most efficient and widely used implementations of GBDT with high performance. We make TSGB publicly available\footnote{Reproducible code for TSGB: https://github.com/felixwzh/TSGB} to encourage further research in tree-based MTL.

\subsection{Evaluation Results}\label{sec:diabetes-exp}
We randomly generate training-validation-testing sets at a ratio of 3:1:1.
The proportion of positive and negative samples can be very different for each task. Therefore the accuracy, recall, and precision are not suitable indicators to measure the performance of models. As we known, AUC (Area Under the Curve of ROC) can be directly compared between tasks with different positive ratios. We take it as the primary indicator and report the average AUC over 10 random seeds in the experiment.
For each algorithm, the best hyperparameters adopted are provided in Appendix~\ref{sec:appendix_A}.

\subsubsection{Multi-center Diabetes Prediction}
The experimental results are presented in Tab.~\ref{tab:diabetes}.
\begin{table}[tbp]
		\centering
		\caption{AUC Scores Under Multi-center Diabetes Dataset.}
		\vspace{-10pt}
        \resizebox{0.47\textwidth}{!}{
		\begin{tabular}{c|ccccccc|c}
            \toprule
			\textbf{Task} & ST-GB & GBDT & MT-ET & MT-TNR & MT-B & CMTL & \textbf{TSGB}$_\lambda$ & \textbf{TSGB}\\
            \midrule
			$1$ & $\textbf{81.04}$ & $80.06$ & $79.82$ & $78.69$ & $78.77$ & $77.85$ & $79.91$ & $\underline{80.65}$\\
            $2$ & $71.62$ & $73.37$ & $71.51$ & $71.81$ & $69.94$ & $71.32$ & $\underline{73.75}$ & $\textbf{73.87}$\\
            $3$ & $77.60$ & $\underline{77.61}$ & $76.20$ & $76.29$ & $74.93$ & $73.50$ & $77.30$ & $\textbf{77.90}$\\
            $4$ & $76.79$ & $77.32$ & $77.62$ & $74.93$ & $76.40$ & $75.07$ & $\underline{77.65}$ & $\textbf{78.18}$\\
            $5$ & $80.91$ & $81.59$ & $80.09$ & $79.25$ & $77.80$ & $77.63$ & $\textbf{81.78}$ & $\underline{81.71}$\\
            $6$ & $80.53$ & $81.32$ & $80.05$ & $78.31$ & $78.72$ & $77.89$ & $\underline{81.38}$ & $\textbf{81.61}$\\
            $7$ & $79.72$ & $80.01$ & $\textbf{82.18}$ & $79.13$ & $77.55$ & $78.58$ & $80.33$ & $\underline{80.50}$\\
            $8$ & $78.25$ & $79.00$ & $74.31$ & $77.51$ & $77.07$ & $76.94$ & $\underline{79.11}$ & $\textbf{79.28}$\\
            $9$ & $77.48$ & $77.54$ & $77.32$ & $76.05$ & $75.67$ & $74.85$ & $\underline{77.79}$ & $\textbf{78.27}$\\
            $10$ & $82.69$ & $\underline{83.21}$ & $83.04$ & $81.89$ & $79.61$ & $80.73$ & $82.96$ & $\textbf{83.31}$\\
            $11$ & $79.34$ & $79.18$ & $\textbf{81.21}$ & $77.93$ & $75.56$ & $76.45$ & $79.40$ & $\underline{79.54}$\\
            $12$ & $72.22$ & $\underline{74.16}$ & $70.29$ & $71.71$ & $72.40$ & $70.48$ & $73.76$ & $\textbf{74.49}$\\
            $13$ & $76.11$ & $78.09$ & $\textbf{80.08}$ & $76.86$ & $75.99$ & $75.37$ & $\underline{78.23}$ & $78.17$\\
            $14$ & $80.44$ & $80.44$ & $77.52$ & $79.35$ & $77.43$ & $78.93$ & $\underline{80.66}$ & $\textbf{80.89}$\\
            $15$ & $86.16$ & $\underline{86.43}$ & $84.02$ & $83.71$ & $84.02$ & $82.79$ & $86.34$ & $\textbf{86.80}$\\
            $16$ & $80.00$ & $79.82$ & $75.24$ & $77.58$ & $75.90$ & $76.51$ & $\underline{80.08}$ & $\textbf{80.11}$\\
            $17$ & $\underline{77.31}$ & $76.84$ & $72.08$ & $74.55$ & $75.58$ & $74.13$ & $77.23$ & $\textbf{77.47}$\\
            $18$ & $60.46$ & $\underline{61.87}$ & $60.01$ & $60.53$ & $60.09$ & $60.46$ & $61.43$ & $\textbf{62.28}$\\
            $19$ & $75.54$ & $74.66$ & $74.42$ & $71.04$ & $73.41$ & $69.26$ & $\textbf{76.44}$ & $\underline{76.17}$\\
            $20$ & $61.75$ & $\underline{63.21}$ & $57.51$ & $61.65$ & $60.65$ & $60.98$ & $63.10$ & $\textbf{63.23}$\\
            $21$ & $82.70$ & $\underline{83.04}$ & $77.23$ & $82.12$ & $79.51$ & $80.59$ & $82.78$ & $\textbf{83.06}$\\
            \midrule
            AVG & $77.08$ & $77.56$ & $75.80$ & $75.76$ & $75.10$ & $74.78$ & $\underline{77.69}$ & $\textbf{77.98}$\\
			\bottomrule
		\end{tabular}}
		\vspace{-10pt}
		\label{tab:diabetes}
	\end{table}
	
The main conclusions can be summarized as follows.
(i) We find ST-GB achieves competitive performance compared to other tree-based MTL models. ST-GB has much better performance than linear MTL models CMTL and MT-TNR.
(ii) GBDT, which trains samples from all the tasks together, boosts 15 tasks' performance compared with ST-GB, but ST-GB still outperforms GBDT on 5 tasks.
This phenomenon is called \textit{negative transfer} \cite{ge2014handling} in MTL, and we owe the main reason to the \textit{negative task gain} problem we analyzed in Sec.~\ref{sec:negtive_task_gain}.
(iii) Although task-wise split is first proposed in MT-ET \cite{simm2014tree}, MT-ET does not achieve satisfactory performance on multi-center diabetes data. The task-wise split criterion in MT-ET is an alternative of one-hot encoding of task feature, which means separate the samples into two random sets of tasks instead of two specific sets version. It is not well designed for the \textit{negative task gain} problem. However, the competitive performance of TSGB$_{\lambda}$ indicates that split the samples task-wisely is promising.
(iv) TSGB outperforms baseline models in almost all the tasks. Specifically, it boosts the performance on 17 of 21 tasks compared to all the other models. TSGB is outperformed by ST-GB on only 1 task with a smaller gap than those between GBDT, MT-ET, TSGB$_{\lambda}$, and ST-GB. This indicates that our analysis of \textit{negative task gain} is reasonable, and our task-wise split mechanism is effective. In conclusion, the results show TSGB is effective on solving \textit{data heterogeneity} and \textit{insufficiency}.

\subsubsection{Multi-domain Sentiment Classification}
The experimental procedures follow the same setting, and we show the main results in Tab.~\ref{tab:amazon}.
\begin{table}[htbp]
	\centering
	\caption{AUC Scores Under Multi-domain Sentiment dataset.}
	\vspace{-10pt}
	\resizebox{0.47\textwidth}{!}{
		\begin{tabular}{c|ccccccc|c}
			\toprule
            \textbf{Task} & ST-GB & GBDT & MT-ET & MT-TNR & MT-B & CMTL & \textbf{TSGB}$_\lambda$ & \textbf{TSGB}\\
            \midrule
            Books & $92.99$ & $93.86$ & $92.79$ & $91.24$ & $93.45$ & $90.83$ & $\underline{93.97}$ & $\textbf{94.37}$\\
            DVDs & $92.57$ & $94.14$ & $92.26$ & $91.15$ & $93.27$ & $90.18$ & $\underline{94.24}$ & $\textbf{94.39}$\\
            Electr. & $95.14$ & $95.98$ & $95.66$ & $93.93$ &  $95.58$ & $93.28$ & $\underline{96.03}$ & $\textbf{96.06}$\\
            K. App. & $96.43$ & $96.99$ & $96.52$ & $94.87$ & $96.74$ & $94.48$ & $\underline{97.11}$ & $\textbf{97.20}$\\
            \midrule
            AVG & $94.28$ & $95.24$ & $94.31$ & $92.80$ & $94.76$ & $92.19$ & $\underline{95.34}$ & $\textbf{95.51}$\\
            \bottomrule
	\end{tabular}}
	\label{tab:amazon}
\end{table}

From AUC scores derived with the whole sentiment dataset, TSGB outperforms all the baseline models. Interestingly, TSGB$_{\lambda}$ reach a good performance second only to TSGB and outperforms MT-ET significantly. We analyze the reasons for this situation is that, although the original task-wise split used in MT-ET boost performance by introducing additional randomness for bagging, such kind of split is an improved realization of encoding task as an additional dimension of feature, and it separates samples into two sets of tasks by a randomly selected task-related value, which can not ensure reducing negative task gain ratio significantly. Different from random many-vs-many split used in MT-ET, we proposed ``ones-vs-rest'' task-wise split in TSGB, which is more targeted for the mentioned negative task gain problem and reduces the negative task gain ratio more effectively. The ones-vs-rest task-wise split means separating tasks with negative task gains and those with positive ones, therefore it is much more reasonable than its original version according to our theoretic analysis and leads to better performance in MTL setting. The analysis also explains that TSGB can perform better than TSGB$_{\lambda}$, since TSGB employs negative task gain ratio as the criterion to perform task-wise split instead of using a constant probability to control whether performing a task-wise split at a certain decision node.

\subsubsection{Robustness to Data Sparsity}\label{sec:robustness_diff_volume}
\begin{figure}[tbp]
\centering
\includegraphics[width=0.55\linewidth]{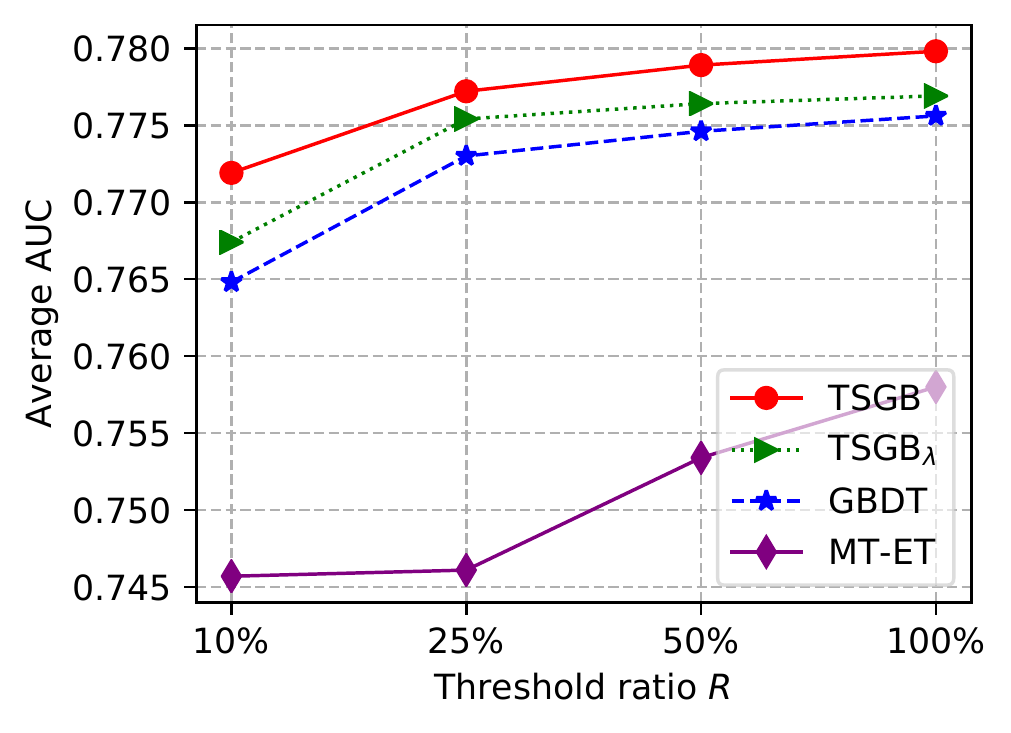}
\vspace{-5pt}
\caption{The performance of TSGB with different training data volume consistently outperforms MT-ET, GBDT, and TSGB$_{\lambda}$.}
\label{fig:auc_percent}
\vspace{-15pt}
\end{figure}

We further study the impact of training data sparsity. We compare TSGB with the best two baselines, TSGB$_{\lambda}$ and GBDT, as well as the original MT-ET on our multi-center diabetes dataset but with different training data volume. We subsample 10\%, 25\%, and 50\% training data on each task and conduct the experiments with the same procedure as before. In Fig.~\ref{fig:auc_percent}, we plot the average AUC of three models on the testing set. It shows that TSGB reaches an average AUC of $0.7772\pm.0005$ with only 25\% training data, while GBDT and TSGB$_{\lambda}$ approach to but still inferior to such an AUC score using 100\% training data (GBDT with average AUC $0.7756\pm.0003$ and TSGB$_{\lambda}$ $0.7769\pm.0005$). MT-ET is the most sensitive to data volume, with performance fluctuates in a large interval.

The observations can be summarized as follows: (i) TSGB reaches a higher average AUC with less training data, which shows that TSGB is robust to data sparsity issue. (ii) Performances of TSGB and TSGB$_{\lambda}$ are far better than the original version of MT-ET. (iii) TSGB outperforms TSGB$_{\lambda}$ and GBDT in most tasks (16$\sim$18 out of 21 tasks) on all the training data volumes. With these three observations, we conclude that task-wise split is helpful in our MTL scenario and conducting task-wise split with consideration of proposed \textit{task gain} further improves the tree-based model's performance.


\begin{figure}[tbp]
\centering
\includegraphics[width=0.45\linewidth]{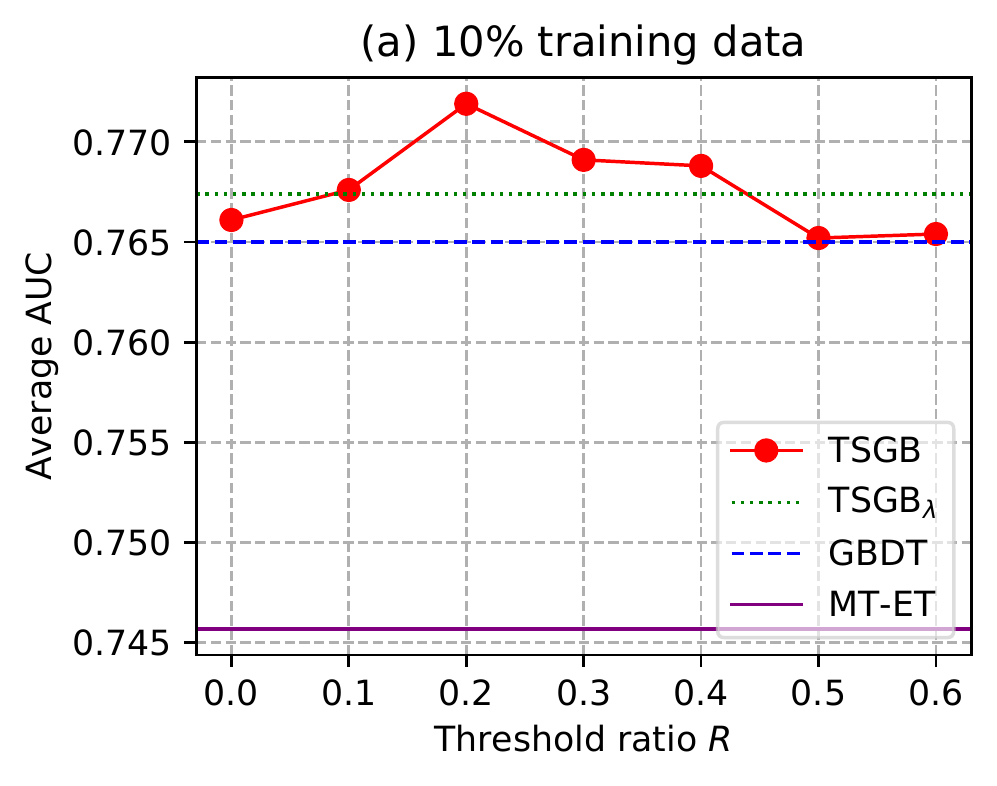}
\includegraphics[width=0.45\linewidth]{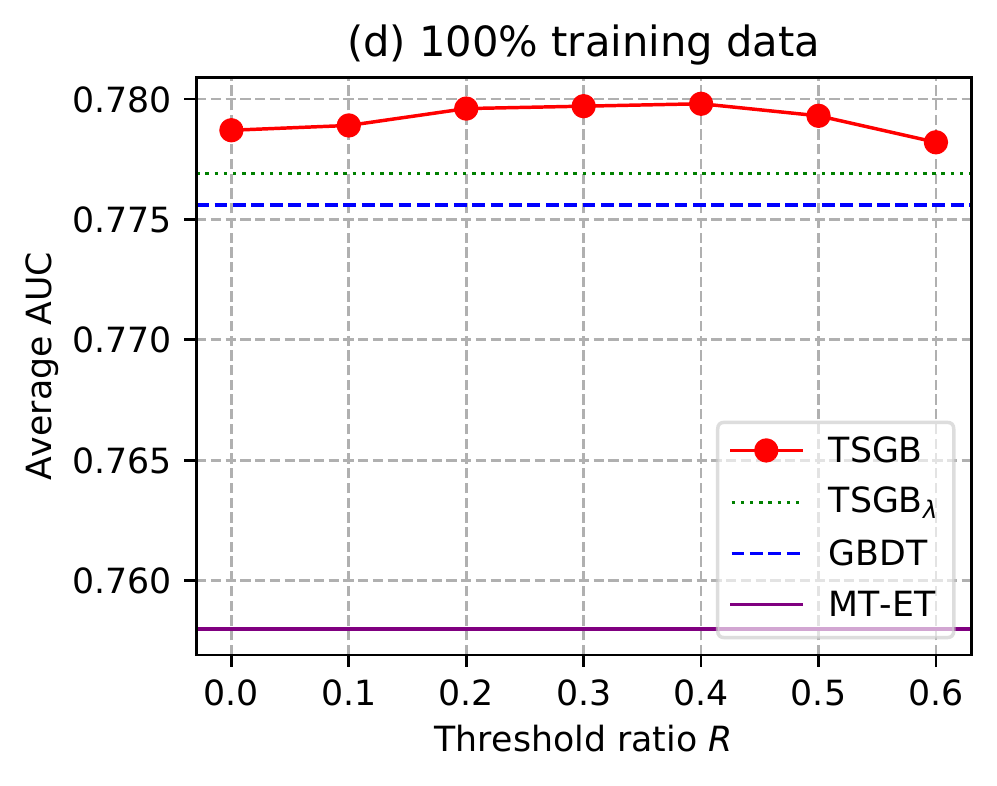}
\vspace{-5pt}
\caption{The performance of TSGB with different threshold ratio $R$ consistently compared with MT-ET, GBDT, and TSGB$_{\lambda}$ on different training data volume.}
\label{fig:ratio_auc}
\vspace{-0.5cm}
\end{figure}

\subsubsection{Hyperparameter Study}\label{sec:param}
We introduce a hyperparameter $R$ as the threshold ratio to determine when to split the node task-wisely. We set $R = 0, 0.1, 0.2, 0.3, 0.4, 0.5, 0.6$ and plot the corresponding average AUC over all the tasks to see the influence of threshold ratio $R$ on TSGB's performance in Fig.~\ref{fig:ratio_auc}. The experimental results at $25\%$ and $50\%$ data volume are very similar to the result at $100\%$ data volume. When the training data is sparse (10\%), the performance difference between TSGB and the other two baselines are small. When there is more training data (25\%, 50\%, and 100\%), TSGB outperforms GBDT and our variant TSGB$_{\lambda}$ consistently. We also find that TSGB has the best performance when $R$ value is set low but not zero, i.e., $R=0.4$ in Fig.~\ref{fig:ratio_auc}(b). If $R$ is too high, we conduct task-wise splits only in a few nodes where the negative task gain problem is severe and fail to handle the problem in many other nodes. On the contrary, if we set the $R$ too low, nearly all the nodes will be split task-wisely (96.47\% nodes, as mentioned in Sec.~\ref{sec:negtive_task_gain}), and only a few nodes can be used to optimize the learning objective. Thus, a relatively low threshold ratio $R\in [0.2,0.4]$ leads to the best performance.


\subsection{Case Study}\label{sec:case_study}
To get deeper insights into how negative task gain problem influences GBDT in MTL, we study one specific task, task-21, with imbalanced down-sampled training data. More specifically, we randomly choose 10\% samples (0.5\% positive \& 99.5\% negative) from task-21, while for the other 20 tasks, we randomly selected 10\% samples (50\% negative \& 50 \% positive) as the training data. One additional reason we build task-21's training data with the very sparse positive sample is that the positive case of some diseases, due to many reasons, might be relatively rare in practice. We want to see whether TSGB could handle this condition and outperform TSGB$_{\lambda}$, GBDT, and ST-GB. In addition, we introduce TSGB-4 to see whether a task-wise split is effective. TSGB-4 means that we use GBDT to train the first three decision trees, but from the fourth tree, we change to TSGB.

\begin{figure}[tbp]
\centering
\includegraphics[width=0.45\linewidth]{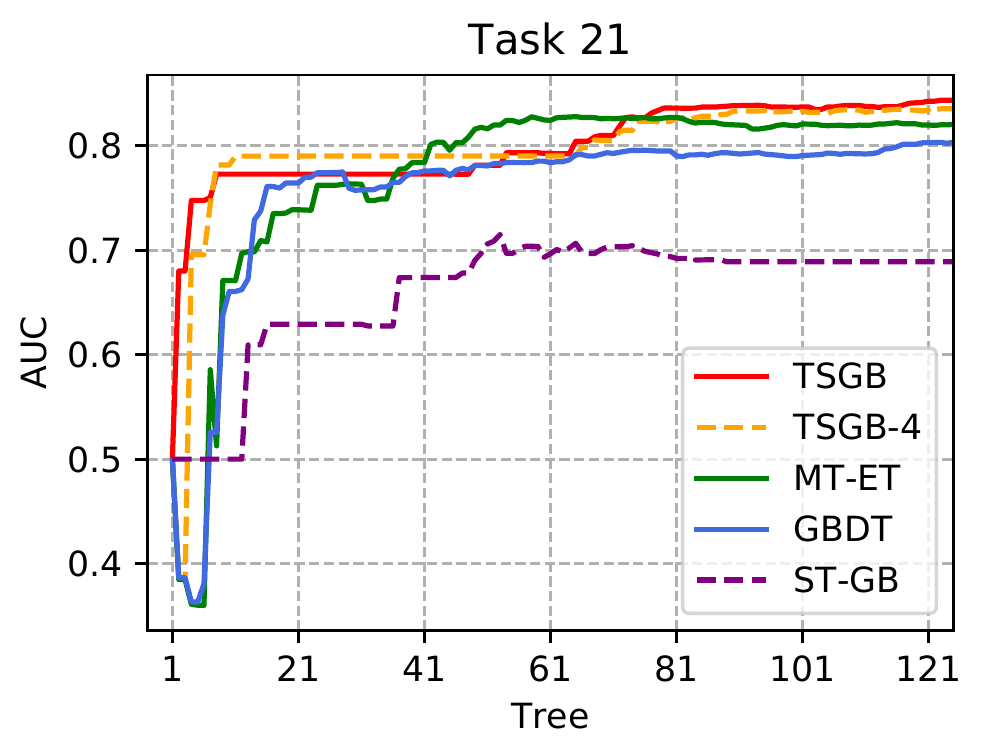}
\includegraphics[width=0.45\linewidth]{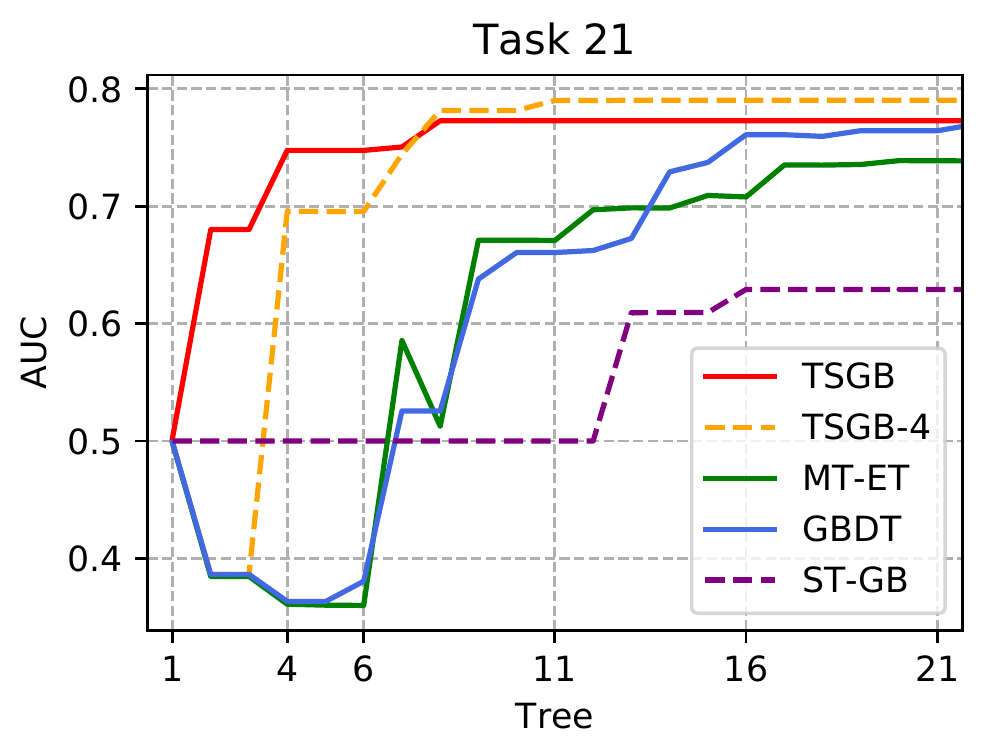}
\vspace{-10pt}
\caption{AUC on task-21's validation set in discussion.}
\label{fig:case-study}
\vspace{-0.5cm}
\end{figure}

From Fig.~\ref{fig:case-study} we can see that (i) when the training data is extremely imbalanced, MTL helps boost model performance. All the MTL models have a large AUC lift compared to ST-GB. (ii) TSGB obtains the highest AUC, which indicates proposed TSGB is capable of better leveraging the training data on all the tasks. (iii) Although directly using GBDT brings about 10\% AUC improvement, the weird AUC curve of GBDT at the first 6 trees (below 0.5, which is worse than the performance of a random classifier) shows some problems of GBDT. (iv) TSGB-4 has exactly the same performance as GBDT in the first three trees, but when the decision tree is constructed in a TSGB from the fourth tree, its performance improves significantly and outperforms GBDT eventually.

We also compare the different behaviors of the fourth decision tree of GBDT and TSGB-4 to see what happened in the training process and have the following findings.
Since we set threshold ratio $R = 0.2$, TSGB-4 will conduct a task-wise split instead of the found best feature-wise split after constructing the fourth tree if $R_{\text{neg}} > 0.2$.
Therefore, with the observation of the negative task gain problem, TSGB converts some feature-wise splits in GBDT into task-wise splits and benefits task-21 from other tasks' training samples. As a result, TSGB-4 boosts AUC on task-21 on the fourth tree and achieves better performance than GBDT when it converges (Fig.~\ref{fig:case-study}).

\section{Application: online diabetes risk assessment software}\label{sec:application}
Rui-Ning Diabetes Risk Assessment is a professional diabetes prediction platform developed by 4Paradigm Inc. It predicts the risk score of healthy people suffering type-2 diabetes in the coming 3 years based on the proposed TSGB. We normalize the model output probability to 1-100 as the risk score.
To make users understand better, we sort the risk score into 4 intervals, 1-30 for good, 31-60 for risk, 61-90 for high-risk, and 91-100 for dangerous.
Beyond that, we also provide key risk factors analysis and corresponding personalized health tips to help the early prevention of type-2 diabetes and guides daily health management.
\begin{figure*}[htbp]
\centering
\includegraphics[width=1\textwidth]{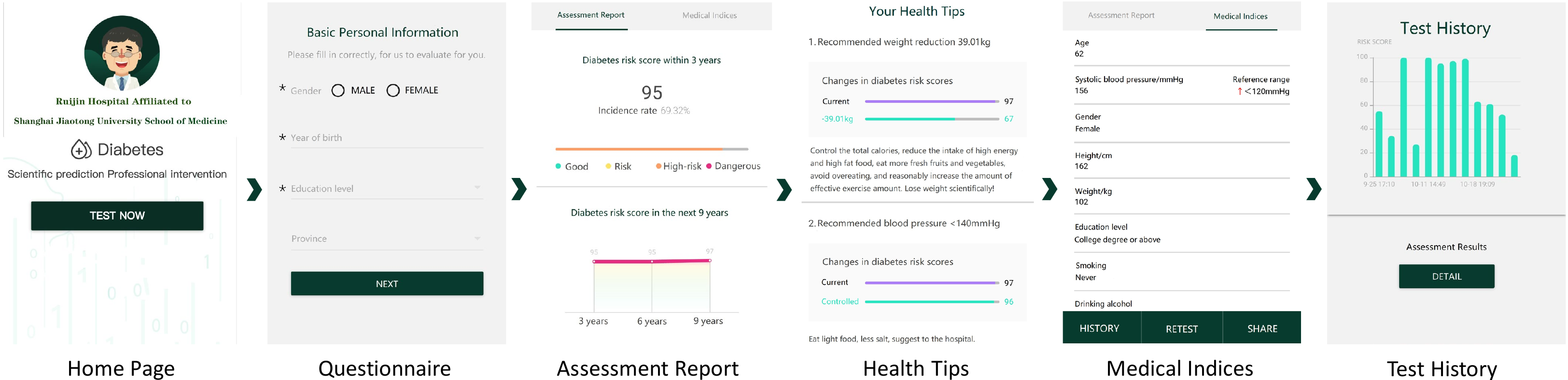}
\vspace{-0.5cm}
\caption{Demonstration of Rui-Ning Diabetes Risk Assessment software workflow.}
\label{fig:ruining}
\vspace{-0.2cm}
\end{figure*}

To start a test, the users need to fill in a questionnaire about their living habits and several medical indicators which can be done within 1 minute (Fig.~\ref{fig:ruining}). We would like to emphasize that, for a rapid prediction, it is impossible to ask users to provide all 50-dimensional features as the training set in practice. Therefore, we select 13 of 50 dimensions, which are the most informative and easy to obtain in medical testing, as the content of the questionnaire (Details in Appendix~\ref{sec:appendix_C}). The characteristics of tree model in dealing with missing features naturally ensure the performance.

In order to evaluate and analyze the performance of Rui-Ning Diabetes Risk Assessment, we employed another 880 healthy volunteers from different regions of China to complete the assessment, then we follow-up visited the volunteers three years later to record whether they get diabetes or not, finally formed the testing data.

\begin{figure}[tbp]
    \centering
    \begin{minipage}[t]{0.45\linewidth}
    \centering
    \includegraphics[width=1\linewidth]{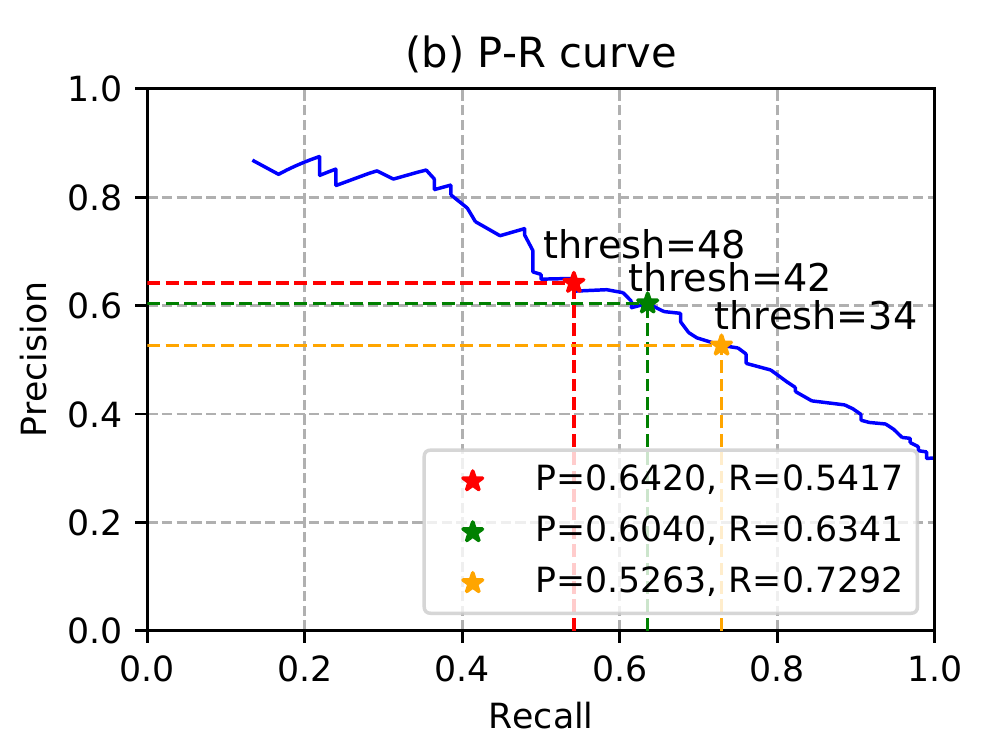}
    \end{minipage}
    \begin{minipage}[t]{0.45\linewidth}
    \centering
    \includegraphics[width=1\linewidth]{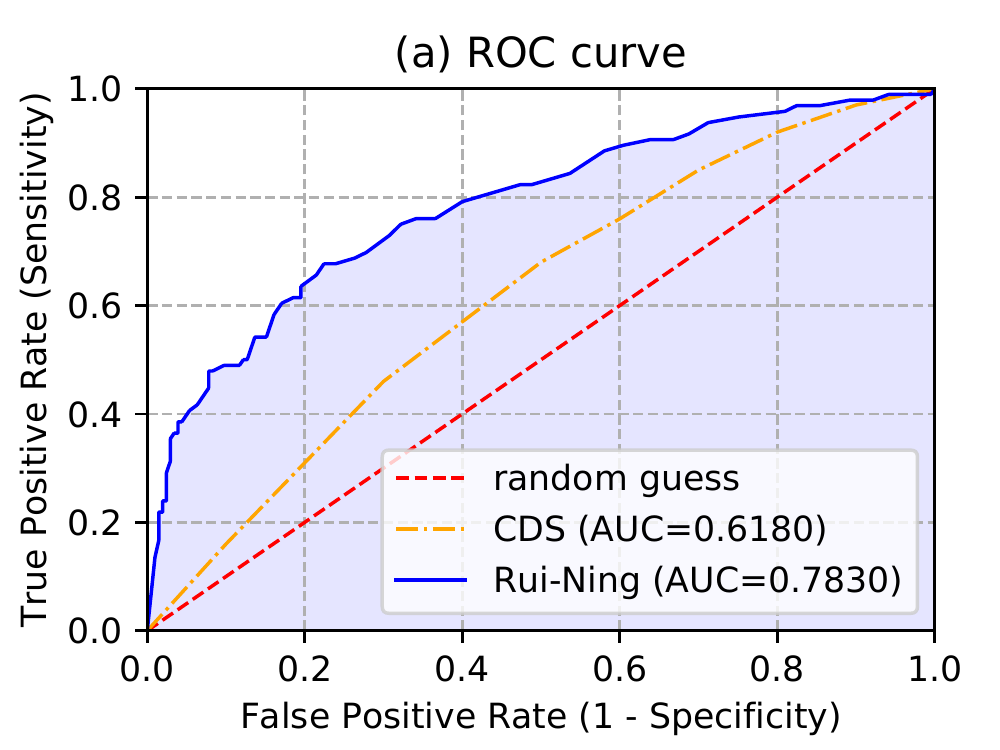}
    \end{minipage}
    \vspace{-5pt}
    \caption{(a) P-R curve for threshold determination, (b) ROC comparison between Rui-Ning and CDS.}
    \vspace{-10pt}
    \label{fig:test_pr/roc}
\end{figure}

As we known, to deploy a binary classification model in a real world scenario, the classification threshold is important. In healthcare domain, the threshold usually be set according to the specific needs. For an example, tumor screening hopes to screen out all positive suspicious, thus tumor screening model focuses on a high sensitivity, which leads high screening costs and relatively low precision. However, in the field of diabetes, it is not the case. For large-scale population, we need to consider the actual economic cost. We must improve the specificity on a certain sensitivity to reduce the actual cost.
To determine the best threshold for Rui-Ning Diabetes Risk Assessment, we plot the P-R curve as in Figure~\ref{fig:test_pr/roc}(a). We can see, when take 42 as the threshold (risk score greater than 42 will be predicted as positive sample), the model has appreciated performance. Based on this threshold, we evaluate the software with multiple indexes as shown in Table~\ref{tab:ruining_eval}.
\begin{table}[htbp]
\vspace{-5pt}
 \caption{Evaluation of deployed software.}
 \vspace{-10pt}
 \resizebox{0.35\textwidth}{!}{
 \begin{tabular}{ccccc}
    \toprule
    Accuracy & Precision & Recall & F1-score & AUC\\
    \midrule
    0.7508 & 0.6040 & 0.6354 & 0.6193 & 0.7830 \\
 \bottomrule
\end{tabular}}
\label{tab:ruining_eval}
\vspace{-5pt}
\end{table}

We then compared the performance of our deployed software with an existing diabetes risk prediction system. To our best knowledge, there are not other open-source softwares that provide diabetes risk assessment in industry, so we employ a traditional rule based diabetes risk scoring method CDS \cite{zhou2013nonlaboratory}, which is a regional authoritative diabetes risk assessment method recommended by Chinese Medical Association, as the main object of comparison.
We plot the ROC curve of our deployment and CDS in Figure~\ref{fig:test_pr/roc}(b).
Rui-Ning significantly improves the AUC from 0.6180 to 0.7963.
The slightly improvement on sensitivity ensures the detection rate of diabetes, while the greatly improvement on the specificity can significantly reduce the cost of screening, which provides a practical and effective prevention and control program in China, a developing country with tight average medical expenses.

\begin{figure}[tbp]
    \centering
    \includegraphics[width=0.45\textwidth]{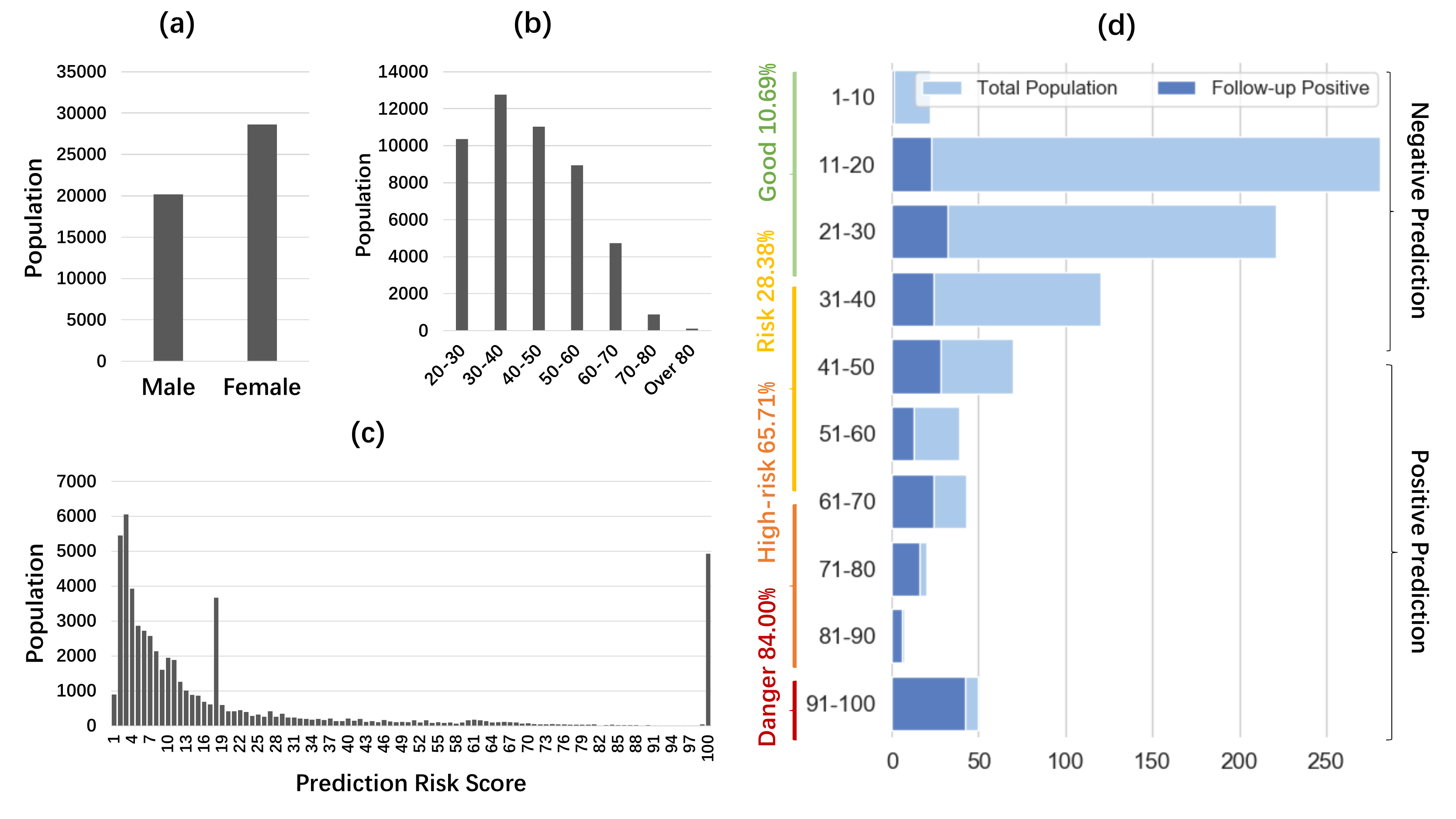}
    \vspace{-10pt}
    \caption{Statistical analysis of user data. (a) Gender distribution, (b) Age distribution, (c) Prediction risk score, (d) Positive rates in different risk groups.}
    \label{fig:ruining_stat}
    \vspace{-0.5cm}
\end{figure}

Rui-Ning Diabetes Risk Assessment aims at providing an efficient and low-cost scheme for huge-scale screening of diabetes, it has been used in different organizations such as physical examination centers, human resource departments and insurance institutes. The software is used by more than $48,000$ people after its deployment. The distribution of all users and their risk scores is illustrated as in Figure~\ref{fig:ruining_stat}(a,b,c). We also discuss the positive rates in different risk groups, the result is drawn in Figure~\ref{fig:ruining_stat}(d). $10.69\%$ of the people were diagnosed with diabetes three years after they were predicted to be in ``Good'' risk group, $28.38\%$ for ``Risk'', $65.71\%$ for ``High-risk'', and $84.00\%$ for ``Dangerous'', which further verified the effectiveness of the deployment software.


\section{Related Works}
\textbf{Multi-Task Learning}.
\textit{Multi-task learning} (MTL) \cite{zhang2017survey} aims to make use of data obtained from multiple related tasks to improve the model performance on all the tasks. MTL helps in many domains under plenty of situations, especially when the amount of data for one learning task is limited, or the cost of collecting data is high.

Consider the types of learning tasks, MTL can be categorized into two sets. Homogeneous MTL deals with learning tasks of the same data space but with different distributions \cite{kumar2012learning}, which is the case discussed in this paper, while heterogeneous MTL processes various types of learning tasks (e.g., classification, regression) \cite{jin2015heterogeneous}.

MTL can also be categorized by the form of knowledge sharing among the tasks. Feature-based MTL learns common features among tasks \cite{argyriou2007multi,maurer2013sparse}, while parameter-based MTL usually leverages the parameter of model from a task to benefit the models from other tasks \cite{ando2005framework,evgeniou2004regularized}.

\vspace{5pt}\noindent\textbf{Tree-based Model in MTL}.
There are a few previous works focused on tree-based models related to MTL. In \cite{goussies2014transfer}, 
a mixed information gain is defined to leverage the knowledge from multiple tasks to train a better learner for a target domain, thus this is not an MTL algorithm but a transfer learning algorithm. The reason why such multiple source domain adaptation algorithms of transfer learning are not suitable for this multi-task diabetes prediction task is that, the setting of transfer learning aims at using knowledge from multiple source domains to improve the performance on target domain, however here we aim at improving performance on every task but not only one task of target domain. Of course we can train several multiple source domain transfer learning models so that improve the performance for every single task, but there is no doubt that such methods are computational expensive and not elegant. An MTL algorithm with boosted decision tree is proposed in \cite{faddoul2012learning}, but it is designed for heterogeneous MTL i.e., the tasks share the same data but with different labels set for each task.

The two previous works which solve the same MTL problem studied in this paper are Multi-task boosting (MT-B) \cite{chapelle2010multi} and Multi-Task ExtraTrees (MT-ET) \cite{simm2014tree}.
The main drawback of MT-B is its computational complexity since it trains $T+1$ forests, and despite being simple and direct, it derives low empirical performance.
In MT-ET, the authors propose to perform a random task split under some predefined probability.
The task split is conducted by sorting the tasks by their ratio of positive samples at this node and find the best split threshold with the highest information gain.
The disadvantage of MT-ET is that the decision of task split is randomly determined and is thus not reasonable since it fails to leverage any task related information, we have verified this through experiments of TSGB$_{\lambda}$.

\section{Conclusion}
In this paper, we proposed the novel Task-wise Split Gradient Boosting Trees (TSGB) model, which extends GBDT to multi-task settings to better leverage data collected from different medical centers. TSGB outperforms several strong baseline models and achieves the best performance in our diabetes prediction task. Moreover, experiments in multi-domain sentiment dataset also show the effectiveness of TSGB in general MTL tasks. 
The discussion further supports our analysis of task gain and negative task gain problem and provides insights of tree-based models in MTL. 
We deployed and productized our online diabetes prediction software Rui-Ning Diabetes Risk Assessment based on proposed TSGB. 
The online software has been widely used by a considerable number of people from different organizations. We have also published our code of TSGB which will help in algorithm reproducibility. Current limitation of TSGB is that when to execute task-wisely is controlled by a predefined hyperparameter of threshold ratio $R$ and needs proper setting.
For future work, we focus on further study the negative transfer problem and hope to explore the possibility of making decision of when to perform task-wise split via reinforcement learning.

\begin{acks}
The project is supported by National Key Research and Development Program of Ministry of Science and Technology of the People's Republic of China (2016YFC1305600, 2018YFC1311800), National Natural Science Foundation of China (82070880, 81771937, 61772333) and Shanghai Municipal Science and Technology Major Project (2021SHZDZX0102).
\end{acks}


\bibliographystyle{ACM-Reference-Format}
\bibliography{sample-base}


\begin{thebibliography}{32}


\ifx \showCODEN    \undefined \def \showCODEN     #1{\unskip}     \fi
\ifx \showDOI      \undefined \def \showDOI       #1{#1}\fi
\ifx \showISBNx    \undefined \def \showISBNx     #1{\unskip}     \fi
\ifx \showISBNxiii \undefined \def \showISBNxiii  #1{\unskip}     \fi
\ifx \showISSN     \undefined \def \showISSN      #1{\unskip}     \fi
\ifx \showLCCN     \undefined \def \showLCCN      #1{\unskip}     \fi
\ifx \shownote     \undefined \def \shownote      #1{#1}          \fi
\ifx \showarticletitle \undefined \def \showarticletitle #1{#1}   \fi
\ifx \showURL      \undefined \def \showURL       {\relax}        \fi
\providecommand\bibfield[2]{#2}
\providecommand\bibinfo[2]{#2}
\providecommand\natexlab[1]{#1}
\providecommand\showeprint[2][]{arXiv:#2}

\bibitem[\protect\citeauthoryear{Ando and Zhang}{Ando and Zhang}{2005}]%
        {ando2005framework}
\bibfield{author}{\bibinfo{person}{Rie~Kubota Ando} {and} \bibinfo{person}{Tong
  Zhang}.} \bibinfo{year}{2005}\natexlab{}.
\newblock \showarticletitle{A framework for learning predictive structures from
  multiple tasks and unlabeled data}.
\newblock \bibinfo{journal}{\emph{JMLR}} (\bibinfo{year}{2005}).
\newblock


\bibitem[\protect\citeauthoryear{Argyriou, Evgeniou, and Pontil}{Argyriou
  et~al\mbox{.}}{2007}]%
        {argyriou2007multi}
\bibfield{author}{\bibinfo{person}{Andreas Argyriou},
  \bibinfo{person}{Theodoros Evgeniou}, {and} \bibinfo{person}{Massimiliano
  Pontil}.} \bibinfo{year}{2007}\natexlab{}.
\newblock \showarticletitle{Multi-task feature learning}. In
  \bibinfo{booktitle}{\emph{NIPS}}.
\newblock


\bibitem[\protect\citeauthoryear{Blitzer, Dredze, and Pereira}{Blitzer
  et~al\mbox{.}}{2007}]%
        {blitzer2007biographies}
\bibfield{author}{\bibinfo{person}{John Blitzer}, \bibinfo{person}{Mark
  Dredze}, {and} \bibinfo{person}{Fernando Pereira}.}
  \bibinfo{year}{2007}\natexlab{}.
\newblock \showarticletitle{Biographies, bollywood, boom-boxes and blenders:
  Domain adaptation for sentiment classification}. In
  \bibinfo{booktitle}{\emph{ACL}}.
\newblock


\bibitem[\protect\citeauthoryear{Breault, Goodall, and Fos}{Breault
  et~al\mbox{.}}{2002}]%
        {breault2002data}
\bibfield{author}{\bibinfo{person}{Joseph~L Breault}, \bibinfo{person}{Colin~R
  Goodall}, {and} \bibinfo{person}{Peter~J Fos}.}
  \bibinfo{year}{2002}\natexlab{}.
\newblock \showarticletitle{Data mining a diabetic data warehouse}.
\newblock \bibinfo{journal}{\emph{Artificial intelligence in medicine}}
  (\bibinfo{year}{2002}).
\newblock


\bibitem[\protect\citeauthoryear{Breiman}{Breiman}{2017}]%
        {breiman2017classification}
\bibfield{author}{\bibinfo{person}{Leo Breiman}.}
  \bibinfo{year}{2017}\natexlab{}.
\newblock \bibinfo{booktitle}{\emph{Classification and regression trees}}.
\newblock \bibinfo{publisher}{Routledge}.
\newblock


\bibitem[\protect\citeauthoryear{Chapelle, Shivaswamy, Vadrevu, Weinberger,
  Zhang, and Tseng}{Chapelle et~al\mbox{.}}{2010}]%
        {chapelle2010multi}
\bibfield{author}{\bibinfo{person}{Olivier Chapelle},
  \bibinfo{person}{Pannagadatta Shivaswamy}, \bibinfo{person}{Srinivas
  Vadrevu}, \bibinfo{person}{Kilian Weinberger}, \bibinfo{person}{Ya Zhang},
  {and} \bibinfo{person}{Belle Tseng}.} \bibinfo{year}{2010}\natexlab{}.
\newblock \showarticletitle{Multi-task learning for boosting with application
  to web search ranking}. In \bibinfo{booktitle}{\emph{KDD}}.
\newblock


\bibitem[\protect\citeauthoryear{Chen, Xu, Weinberger, and Sha}{Chen
  et~al\mbox{.}}{2012}]%
        {chen2012marginalized}
\bibfield{author}{\bibinfo{person}{Minmin Chen}, \bibinfo{person}{Zhixiang Xu},
  \bibinfo{person}{Kilian Weinberger}, {and} \bibinfo{person}{Fei Sha}.}
  \bibinfo{year}{2012}\natexlab{}.
\newblock \showarticletitle{Marginalized denoising autoencoders for domain
  adaptation}.
\newblock \bibinfo{journal}{\emph{arXiv}} (\bibinfo{year}{2012}).
\newblock


\bibitem[\protect\citeauthoryear{Chen and Guestrin}{Chen and Guestrin}{2016}]%
        {chen2016xgboost}
\bibfield{author}{\bibinfo{person}{Tianqi Chen} {and} \bibinfo{person}{Carlos
  Guestrin}.} \bibinfo{year}{2016}\natexlab{}.
\newblock \showarticletitle{Xgboost: A scalable tree boosting system}. In
  \bibinfo{booktitle}{\emph{KDD}}.
\newblock


\bibitem[\protect\citeauthoryear{Evgeniou and Pontil}{Evgeniou and
  Pontil}{2004}]%
        {evgeniou2004regularized}
\bibfield{author}{\bibinfo{person}{Theodoros Evgeniou} {and}
  \bibinfo{person}{Massimiliano Pontil}.} \bibinfo{year}{2004}\natexlab{}.
\newblock \showarticletitle{Regularized multi--task learning}. In
  \bibinfo{booktitle}{\emph{KDD}}.
\newblock


\bibitem[\protect\citeauthoryear{Faddoul, Chidlovskii, Gilleron, and
  Torre}{Faddoul et~al\mbox{.}}{2012}]%
        {faddoul2012learning}
\bibfield{author}{\bibinfo{person}{Jean~Baptiste Faddoul},
  \bibinfo{person}{Boris Chidlovskii}, \bibinfo{person}{R{\'e}mi Gilleron},
  {and} \bibinfo{person}{Fabien Torre}.} \bibinfo{year}{2012}\natexlab{}.
\newblock \showarticletitle{Learning multiple tasks with boosted decision
  trees}. In \bibinfo{booktitle}{\emph{ECML PKDD}}.
\newblock


\bibitem[\protect\citeauthoryear{Friedman}{Friedman}{2001}]%
        {friedman2001greedy}
\bibfield{author}{\bibinfo{person}{Jerome~H Friedman}.}
  \bibinfo{year}{2001}\natexlab{}.
\newblock \showarticletitle{Greedy function approximation: a gradient boosting
  machine}.
\newblock \bibinfo{journal}{\emph{Annals of statistics}}
  (\bibinfo{year}{2001}).
\newblock


\bibitem[\protect\citeauthoryear{Friedman}{Friedman}{2002}]%
        {friedman2002stochastic}
\bibfield{author}{\bibinfo{person}{Jerome~H Friedman}.}
  \bibinfo{year}{2002}\natexlab{}.
\newblock \showarticletitle{Stochastic gradient boosting}.
\newblock \bibinfo{journal}{\emph{CSDA}} (\bibinfo{year}{2002}).
\newblock


\bibitem[\protect\citeauthoryear{Ge, Gao, Ngo, Li, and Zhang}{Ge
  et~al\mbox{.}}{2014}]%
        {ge2014handling}
\bibfield{author}{\bibinfo{person}{Liang Ge}, \bibinfo{person}{Jing Gao},
  \bibinfo{person}{Hung Ngo}, \bibinfo{person}{Kang Li}, {and}
  \bibinfo{person}{Aidong Zhang}.} \bibinfo{year}{2014}\natexlab{}.
\newblock \showarticletitle{On handling negative transfer and imbalanced
  distributions in multiple source transfer learning}.
\newblock \bibinfo{journal}{\emph{SADM}} (\bibinfo{year}{2014}).
\newblock


\bibitem[\protect\citeauthoryear{Goodfellow, Bengio, Courville, and
  Bengio}{Goodfellow et~al\mbox{.}}{2016}]%
        {goodfellow2016deep}
\bibfield{author}{\bibinfo{person}{Ian Goodfellow}, \bibinfo{person}{Yoshua
  Bengio}, \bibinfo{person}{Aaron Courville}, {and} \bibinfo{person}{Yoshua
  Bengio}.} \bibinfo{year}{2016}\natexlab{}.
\newblock \bibinfo{booktitle}{\emph{Deep learning}}. Vol.~\bibinfo{volume}{1}.
\newblock \bibinfo{publisher}{MIT press Cambridge}.
\newblock


\bibitem[\protect\citeauthoryear{Goodman, Lessler, Cosgrove, Harris,
  Lautenbach, Han, Milstone, Massey, and Tamma}{Goodman et~al\mbox{.}}{2016}]%
        {goodman2016clinical}
\bibfield{author}{\bibinfo{person}{Katherine~E Goodman},
  \bibinfo{person}{Justin Lessler}, \bibinfo{person}{Sara~E Cosgrove},
  \bibinfo{person}{Anthony~D Harris}, \bibinfo{person}{Ebbing Lautenbach},
  \bibinfo{person}{Jennifer~H Han}, \bibinfo{person}{Aaron~M Milstone},
  \bibinfo{person}{Colin~J Massey}, {and} \bibinfo{person}{Pranita~D Tamma}.}
  \bibinfo{year}{2016}\natexlab{}.
\newblock \showarticletitle{A clinical decision tree to predict whether a
  bacteremic patient is infected with an extended-spectrum
  $\beta$-lactamase--producing organism}.
\newblock \bibinfo{journal}{\emph{CID}} (\bibinfo{year}{2016}).
\newblock


\bibitem[\protect\citeauthoryear{Goussies, Ubalde, and Mejail}{Goussies
  et~al\mbox{.}}{2014}]%
        {goussies2014transfer}
\bibfield{author}{\bibinfo{person}{Norberto~A Goussies},
  \bibinfo{person}{Sebasti{\'a}n Ubalde}, {and} \bibinfo{person}{Marta
  Mejail}.} \bibinfo{year}{2014}\natexlab{}.
\newblock \showarticletitle{Transfer learning decision forests for gesture
  recognition}.
\newblock \bibinfo{journal}{\emph{JMLR}} (\bibinfo{year}{2014}).
\newblock


\bibitem[\protect\citeauthoryear{Ioffe and Szegedy}{Ioffe and Szegedy}{2015}]%
        {ioffe2015batch}
\bibfield{author}{\bibinfo{person}{Sergey Ioffe} {and}
  \bibinfo{person}{Christian Szegedy}.} \bibinfo{year}{2015}\natexlab{}.
\newblock \showarticletitle{Batch normalization: Accelerating deep network
  training by reducing internal covariate shift}.
\newblock \bibinfo{journal}{\emph{arXiv}} (\bibinfo{year}{2015}).
\newblock


\bibitem[\protect\citeauthoryear{Ji and Ye}{Ji and Ye}{2009}]%
        {ji2009accelerated}
\bibfield{author}{\bibinfo{person}{Shuiwang Ji} {and} \bibinfo{person}{Jieping
  Ye}.} \bibinfo{year}{2009}\natexlab{}.
\newblock \showarticletitle{An accelerated gradient method for trace norm
  minimization}. In \bibinfo{booktitle}{\emph{ICML}}.
\newblock


\bibitem[\protect\citeauthoryear{Jin, Zhuang, Pan, Du, Luo, and He}{Jin
  et~al\mbox{.}}{2015}]%
        {jin2015heterogeneous}
\bibfield{author}{\bibinfo{person}{Xin Jin}, \bibinfo{person}{Fuzhen Zhuang},
  \bibinfo{person}{Sinno~Jialin Pan}, \bibinfo{person}{Changying Du},
  \bibinfo{person}{Ping Luo}, {and} \bibinfo{person}{Qing He}.}
  \bibinfo{year}{2015}\natexlab{}.
\newblock \showarticletitle{Heterogeneous multi-task semantic feature learning
  for classification}. In \bibinfo{booktitle}{\emph{CIKM}}.
\newblock


\bibitem[\protect\citeauthoryear{Johnson and Zhang}{Johnson and Zhang}{2014}]%
        {johnson2014learning}
\bibfield{author}{\bibinfo{person}{Rie Johnson} {and} \bibinfo{person}{Tong
  Zhang}.} \bibinfo{year}{2014}\natexlab{}.
\newblock \showarticletitle{Learning nonlinear functions using regularized
  greedy forest}.
\newblock \bibinfo{journal}{\emph{TPAMI}} (\bibinfo{year}{2014}).
\newblock


\bibitem[\protect\citeauthoryear{Koh, Tan, et~al\mbox{.}}{Koh
  et~al\mbox{.}}{2011}]%
        {koh2011data}
\bibfield{author}{\bibinfo{person}{Hian~Chye Koh}, \bibinfo{person}{Gerald
  Tan}, {et~al\mbox{.}}} \bibinfo{year}{2011}\natexlab{}.
\newblock \showarticletitle{Data mining applications in healthcare}.
\newblock \bibinfo{journal}{\emph{JHIM}} (\bibinfo{year}{2011}).
\newblock


\bibitem[\protect\citeauthoryear{Kumar and Daum{\'e}~III}{Kumar and
  Daum{\'e}~III}{2012}]%
        {kumar2012learning}
\bibfield{author}{\bibinfo{person}{Abhishek Kumar} {and} \bibinfo{person}{Hal
  Daum{\'e}~III}.} \bibinfo{year}{2012}\natexlab{}.
\newblock \showarticletitle{Learning task grouping and overlap in multi-task
  learning}. In \bibinfo{booktitle}{\emph{ICML}}.
\newblock


\bibitem[\protect\citeauthoryear{Maurer, Pontil, and Romera-Paredes}{Maurer
  et~al\mbox{.}}{2013}]%
        {maurer2013sparse}
\bibfield{author}{\bibinfo{person}{Andreas Maurer}, \bibinfo{person}{Massi
  Pontil}, {and} \bibinfo{person}{Bernardino Romera-Paredes}.}
  \bibinfo{year}{2013}\natexlab{}.
\newblock \showarticletitle{Sparse coding for multitask and transfer learning}.
  In \bibinfo{booktitle}{\emph{ICML}}.
\newblock


\bibitem[\protect\citeauthoryear{Pedregosa, Varoquaux, Gramfort, Michel,
  Thirion, Grisel, Blondel, Prettenhofer, Weiss, Dubourg,
  et~al\mbox{.}}{Pedregosa et~al\mbox{.}}{2011}]%
        {pedregosa2011scikit}
\bibfield{author}{\bibinfo{person}{Fabian Pedregosa}, \bibinfo{person}{Ga{\"e}l
  Varoquaux}, \bibinfo{person}{Alexandre Gramfort}, \bibinfo{person}{Vincent
  Michel}, \bibinfo{person}{Bertrand Thirion}, \bibinfo{person}{Olivier
  Grisel}, \bibinfo{person}{Mathieu Blondel}, \bibinfo{person}{Peter
  Prettenhofer}, \bibinfo{person}{Ron Weiss}, \bibinfo{person}{Vincent
  Dubourg}, {et~al\mbox{.}}} \bibinfo{year}{2011}\natexlab{}.
\newblock \showarticletitle{Scikit-learn: Machine learning in Python}.
\newblock \bibinfo{journal}{\emph{JMLR}} (\bibinfo{year}{2011}).
\newblock


\bibitem[\protect\citeauthoryear{Qin, Xia, and Li}{Qin et~al\mbox{.}}{2009}]%
        {qin2009dtu}
\bibfield{author}{\bibinfo{person}{Biao Qin}, \bibinfo{person}{Yuni Xia}, {and}
  \bibinfo{person}{Fang Li}.} \bibinfo{year}{2009}\natexlab{}.
\newblock \showarticletitle{DTU: a decision tree for uncertain data}. In
  \bibinfo{booktitle}{\emph{PAKDD}}.
\newblock


\bibitem[\protect\citeauthoryear{Qu, Fang, Zhang, Tang, Niu, Guo, Yu, and
  He}{Qu et~al\mbox{.}}{2018}]%
        {qu2018product}
\bibfield{author}{\bibinfo{person}{Yanru Qu}, \bibinfo{person}{Bohui Fang},
  \bibinfo{person}{Weinan Zhang}, \bibinfo{person}{Ruiming Tang},
  \bibinfo{person}{Minzhe Niu}, \bibinfo{person}{Huifeng Guo},
  \bibinfo{person}{Yong Yu}, {and} \bibinfo{person}{Xiuqiang He}.}
  \bibinfo{year}{2018}\natexlab{}.
\newblock \showarticletitle{Product-based Neural Networks for User Response
  Prediction over Multi-field Categorical Data}.
\newblock \bibinfo{journal}{\emph{arXiv}} (\bibinfo{year}{2018}).
\newblock


\bibitem[\protect\citeauthoryear{Ridgeway}{Ridgeway}{2007}]%
        {ridgeway2007generalized}
\bibfield{author}{\bibinfo{person}{Greg Ridgeway}.}
  \bibinfo{year}{2007}\natexlab{}.
\newblock \showarticletitle{Generalized Boosted Models: A guide to the gbm
  package}.
\newblock \bibinfo{journal}{\emph{Update}} (\bibinfo{year}{2007}).
\newblock


\bibitem[\protect\citeauthoryear{Simm, de~Abril, and Sugiyama}{Simm
  et~al\mbox{.}}{2014}]%
        {simm2014tree}
\bibfield{author}{\bibinfo{person}{Jaak Simm},
  \bibinfo{person}{Ildefons~Magrans de Abril}, {and} \bibinfo{person}{Masashi
  Sugiyama}.} \bibinfo{year}{2014}\natexlab{}.
\newblock \showarticletitle{Tree-based ensemble multi-task learning method for
  classification and regression}.
\newblock \bibinfo{journal}{\emph{IEICE TRANSACTIONS on Information and
  Systems}} (\bibinfo{year}{2014}).
\newblock


\bibitem[\protect\citeauthoryear{Walker, Mehalick, Glueck, Tschiffely,
  Cunningham, Norris, and Davidson}{Walker et~al\mbox{.}}{2017}]%
        {walker2017decision}
\bibfield{author}{\bibinfo{person}{Peter~B Walker}, \bibinfo{person}{Melissa~L
  Mehalick}, \bibinfo{person}{Amanda~C Glueck}, \bibinfo{person}{Anna~E
  Tschiffely}, \bibinfo{person}{Craig~A Cunningham}, \bibinfo{person}{Jacob~N
  Norris}, {and} \bibinfo{person}{Ian~N Davidson}.}
  \bibinfo{year}{2017}\natexlab{}.
\newblock \showarticletitle{A decision tree framework for understanding
  blast-induced mild Traumatic Brain Injury in a military medical database}.
\newblock \bibinfo{journal}{\emph{JDMS}} (\bibinfo{year}{2017}).
\newblock


\bibitem[\protect\citeauthoryear{Zhang and Yang}{Zhang and Yang}{2017}]%
        {zhang2017survey}
\bibfield{author}{\bibinfo{person}{Yu Zhang} {and} \bibinfo{person}{Qiang
  Yang}.} \bibinfo{year}{2017}\natexlab{}.
\newblock \showarticletitle{A survey on multi-task learning}.
\newblock \bibinfo{journal}{\emph{arXiv}} (\bibinfo{year}{2017}).
\newblock


\bibitem[\protect\citeauthoryear{Zhou, Chen, and Ye}{Zhou
  et~al\mbox{.}}{2011}]%
        {zhou2011clustered}
\bibfield{author}{\bibinfo{person}{Jiayu Zhou}, \bibinfo{person}{Jianhui Chen},
  {and} \bibinfo{person}{Jieping Ye}.} \bibinfo{year}{2011}\natexlab{}.
\newblock \showarticletitle{Clustered multi-task learning via alternating
  structure optimization}. In \bibinfo{booktitle}{\emph{NIPS}}.
\newblock


\bibitem[\protect\citeauthoryear{Zhou, Qiao, Ji, Ning, Yang, Weng, Shan, Tian,
  Ji, Lin, et~al\mbox{.}}{Zhou et~al\mbox{.}}{2013}]%
        {zhou2013nonlaboratory}
\bibfield{author}{\bibinfo{person}{Xianghai Zhou}, \bibinfo{person}{Qing Qiao},
  \bibinfo{person}{Linong Ji}, \bibinfo{person}{Feng Ning},
  \bibinfo{person}{Wenying Yang}, \bibinfo{person}{Jianping Weng},
  \bibinfo{person}{Zhongyan Shan}, \bibinfo{person}{Haoming Tian},
  \bibinfo{person}{Qiuhe Ji}, \bibinfo{person}{Lixiang Lin}, {et~al\mbox{.}}}
  \bibinfo{year}{2013}\natexlab{}.
\newblock \showarticletitle{Nonlaboratory-based risk assessment algorithm for
  undiagnosed type 2 diabetes developed on a nation-wide diabetes survey}.
\newblock \bibinfo{journal}{\emph{Diabetes care}} \bibinfo{volume}{36},
  \bibinfo{number}{12} (\bibinfo{year}{2013}), \bibinfo{pages}{3944--3952}.
\newblock


\end{thebibliography}


\appendix
\section{Hyperparameters Settings}\label{sec:appendix_A}
In this section, we present the search spaces of hyperparameters in our experiments, as well as the details of hyperparameters settings we finally used, which helps in algorithm reproducing.

In our experiments, we firstly fix the training hyperparameters and consider different combinations of tree booster hyperparameters. Specifically, we search maximum tree depth from $\{3, 4, 5, 6, 7, 8, 9\}$, minimum leaf node sample weight sum from $\{1, 3, 5, 7, 9, 11\}$, sample rates from $[0.6, 1.0)$, and $\gamma$ from $[0.05, 0.45]$. After effective tree booster hyperparameters are found, we further search learning rate $\eta$ from $[0.1, 0.6]$ and regularization weight $\alpha$ from $\{1\times 10^{-5}, 5\times 10^{-4}, 1\times 10^{-3}, 0.1, 1\}$. We finally finetune the hyperparameters around the current optimum, and determine the threshold ratio $R$ following Sec.~\ref{sec:param}. The best hyperparameters for the diabetes dataset and sentiment dataset are listed in Tab.~\ref{tab:param_mtgbdt}.
\begin{table}[htbp]
 \caption{The TSGB hyperparameters on different datasets}
 \vspace{-0.4cm}
 \resizebox{0.35\textwidth}{!}{
 \begin{tabular}{lll}
    \toprule
    \textbf{Hyperparameter}&Diabetes&Sentiment\\
    \midrule
    max\_depth              & 5   & 9\\
    min\_child\_weight      & 5   & 1\\
    colsample\_bytree       & 0.7 & 1.0\\
    colsample\_bylevel      & 0.8 & 0.8\\
    subsample               & 0.8 & 1.0\\
    gamma                   & 0.2 & 0.45\\
    learning
    \_rate          & 0.1 & 0.3\\
    reg\_alpha              & 0.1 & 0.0005\\
    reg\_lambda             & 12  & 12\\
    max\_neg\_sample\_ratio & 0.4 & 0.5\\
 \bottomrule
\end{tabular}}
\label{tab:param_mtgbdt}
\end{table}

We use the optimal parameters of different models and randomly selected 10 initial seeds to run each training-evaluation process, and calculate the average AUC score as the final result. The 95\% confidence intervals is given by $Z_{.025}\times \sigma / \sqrt{n}$, where $Z_{.025} = 1.96$, $n = 10$, $\sigma$ is standard deviation.
In this experimental settings, the more detailed corresponding results are shown in Tab.~\ref{tab:21tasks-10-25-50-100} and Tab.~\ref{tab:4tasks-10-25-50-100}. 

\section{Experimental Settings}\label{sec:appendix_B}
In this section, we provide detailed experimental settings of Sec.~\ref{sec:case_study} for better reproducibility of such kind of case study.

\begin{figure*}
\centering
\includegraphics[width=0.985\linewidth]{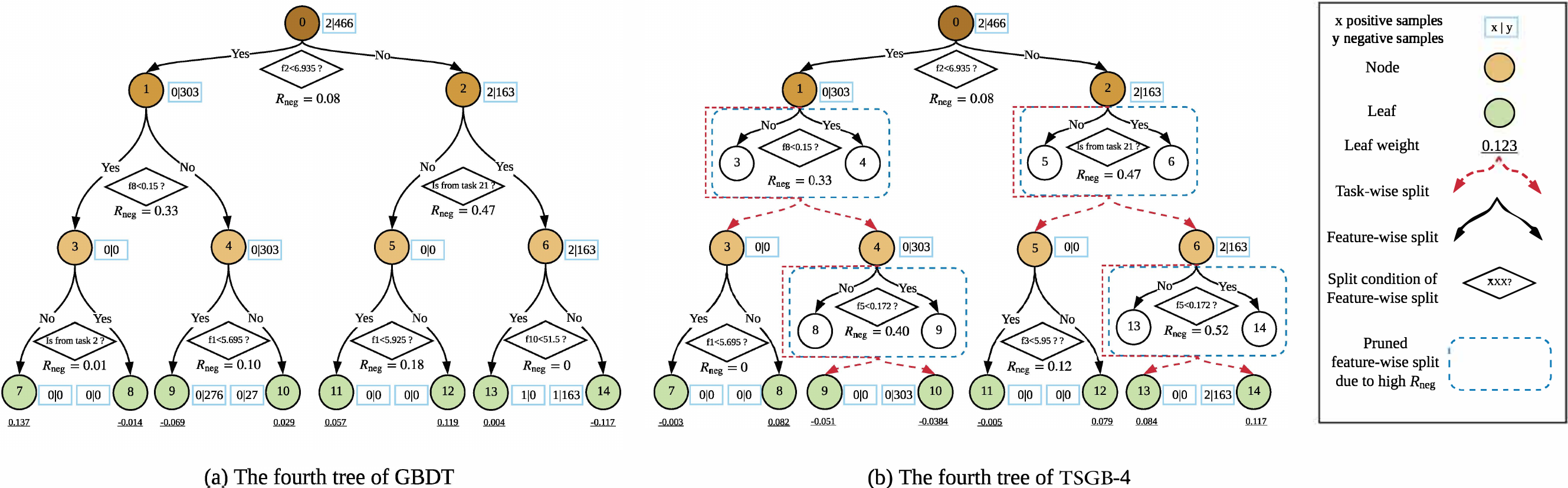}
\caption{An example of visualization of the fourth decision tree of GBDT and TSGB-4 in case study, where f1, f2, f3, f5, f8 and f10 are all code names of the features in our diabetes dataset. Corresponding details of task-wise splits at node-1,2,4,6 in TSGB-4 refer to in Appendix~\ref{sec:appendix_B} Tab.~\ref{tab:node-task}. }
\label{fig:case_study_mt_gbdt}
\end{figure*}

We set threshold ratio $R=0.2$, and then train GBDT and TSGB-4 on task-21 dataset respectively. Therefore, how each task's samples go through the tree in detail should be like that shown in Tab.~\ref{tab:node-task}. Take a vivid example, once we plot a certain decision tree (the fourth here) of GBDT and TSGB-4, we can get a view similar to Fig.~\ref{fig:case_study_mt_gbdt}. The (positive sample number $|$ negative sample number) pairs of task-21 in each node and leaf show how task-21's samples go through the decision tree. The $R_{\text{neg}}$ defined in Eq.~\eqref{eq:R_neg} is provided with the corresponding split condition at each node. 
\begin{table}[htbp]
	\centering
	\caption{An example of how each task's samples go through the tree shown in Fig.~\ref{fig:case_study_mt_gbdt}.}
	\resizebox{0.45\textwidth}{!}{
		\begin{tabular}{c|c|c}
			\toprule
			\textbf{Node} & TSGB-4 & GBDT\\
            \midrule
            $1$ & $\mathcal{A}$ & $\mathcal{A}$\\
            $2$ & $\mathcal{A}$ & $\mathcal{A}$\\
            $3$ & $\{1,2,3,4,6,9,17\}$ & $\{2,4,11,12,17,18,20\}$\\
            $4$ & $\mathcal{A}\setminus\{1,2,3,4,6,9,17\}$ & $\mathcal{A}\setminus\{18,20\}$\\
            $5$ & $\{5,7,8,9,10,11,13,16,19\}$ & $\mathcal{A}\setminus\{21\}$\\
            $6$ & $\{1,2,3,4,6,12,14,15,17,18,20,21\}$ & $\{21\}$\\
            $7$ & $\{1,2,3,4,6,9,17\}$ & $\{4,11,12,17,18,20\}$\\
            $8$ & $\{1,2,3,4,6,9,17\}$ & $\{2\}$\\
            $9$ & $\{8,10,14,15,16,19\}$ & $\mathcal{A}\setminus\{18,20 \}$\\
            $10$ & $\{5,7,11,12,13,18,20,21\}$ & $\mathcal{A}\setminus\{18,20\}$\\
            $11$ & $\{5,7,8,9,10,11,13,16,19\}$ & $\mathcal{A}\setminus\{21\}$\\
            $12$ & $\{5,7,8,9,10,11,13,16,19\}$ & $\mathcal{A}\setminus\{21\}$\\
            $13$ & $\{1,2,4,14,15,17\}$ & $\{21\}$\\
            $14$ & $\{3,6,12,18,20,21\}$ & $\{21\}$\\
			\bottomrule
	\end{tabular}}
	\label{tab:node-task}
    \footnotesize \flushleft{Note: $\mathcal{A}=\{x|x\in\mathbb{N},1\leq x \leq 21 \}$ is the set of all the 21 tasks.}
    \vspace{-10pt}
\end{table}

In GBDT (Fig.~\ref{fig:case_study_mt_gbdt}(a)), when 2 positive and 163 negative samples from task-21 are assigned to node-2, the best split condition in GBDT is dividing the samples based on whether a sample is from task-21. This is because most samples of task-21 in node-2 are negative samples, and such a split condition could minimize the overall objective than others. But this condition is not good to task-21 since only task-21's samples could go through the sub-tree ($2\to6\to(13,14)$), this sub-tree is actually constructed in a single-task manner, and task-21 could not benefit from other tasks' data. This sub-tree structure is also related to the AUC below 0.5 on task-21, because the model is biased by task-21's imbalanced label distribution, and almost all of the samples on the right branch from node-2 will be given a negative predicted value.

In TSGB-4 (Fig.~\ref{fig:case_study_mt_gbdt}(b)), the first split at root node-0 is the same with GBDT, as the optimal split condition ``f2<6.935?'' only has $R_{\text{neg}}=0.08$ and is thus good to most of the tasks. But when it comes to node-2 on the right branch, the optimal split condition ``Is from task 21?'' will actually increase, instead of reduce,
near half of the tasks will have negative task gain problem. In this condition, TSGB will instead split the samples task-wisely, with tasks with negative task gain to the left branch and positive ones to the right. 
Although the same 2 positive and 163 negative samples of task-21 are assigned to node-6 as GBDT in Fig.~\ref{fig:case_study_mt_gbdt}(b), there also are 2431 samples from other 11 tasks in node-6, which means task-21 could benefit from other tasks by sharing more tree structure. At node-6, the optimal split condition, again, leads to negative task gain problem in about half the samples, which means for more than the half samples, they had better not split following ``f5<0.172?''. Therefore, TSGB conducts task-wise split at node-6 and assigns task-21's samples to leaf-14 with other 1088 samples from 5 tasks. One may doubt the task-wise split on the last level in a decision tree. Actually, the rationale lies in the decision tree's nature of dichotomy, and therefore task-wise split is an alternative to feature-wise split to find a generally ``good'' division when feature-wise split cannot.


\section{Application Settings}\label{sec:appendix_C}
The deployment environment is based on servers with 2 regular nodes and 2 test nodes. Each node is in CentOS Linux release 7.6.1810 with 4 cores 2394MHz CPU, 32GB RAM, and 50GB SSD.

Users are required to provide a series of basic personal information and living habits containing gender, year of birth, educational background, height, weight, family history of diabetes, hypertension, fatty liver, smoking and drinking. As an optional item, user can also provide province, fasting blood glucose and systolic blood pressure for a more accurate diabetes risk assessment.


\begin{table*}
	\centering
	\caption{Complete AUC results on diabetes dataset for hyperparameters listed in Tab.~\ref{tab:param_mtgbdt}.}
	\label{tab:21tasks-10-25-50-100}
	\resizebox{0.985\textwidth}{!}{
		\begin{tabular}{c|cccc|cccc|cccc}
			\toprule
			\multirow{2}{*}{\textbf{Task}} & \multicolumn{4}{c|}{$\textbf{25\%}$} & \multicolumn{4}{c|}{$\textbf{50\%}$} & \multicolumn{4}{c}{$\textbf{100\%}$} \\
			& ~GBDT~~ & ~MT-ET~ & TSGB$_{\lambda}$ & TSGB & ~GBDT~~ & ~MT-ET~ & TSGB$_{\lambda}$ & TSGB & ~GBDT~~ & ~MT-ET~ & TSGB$_{\lambda}$ & TSGB\\
			\midrule
			$1$ & $\underline{79.89}{\pm.20}$ & $77.22{\pm.24}$ & $79.86{\pm.21}$ & $\textbf{80.45}{\pm.36}$ & $79.91{\pm.15}$ & $77.79{\pm.15}$ & $\underline{79.98}{\pm.17}$ & $\textbf{80.54}{\pm.37}$ & $\underline{80.06}{\pm.20}$ & $79.82{\pm.19}$ & $79.91{\pm.12}$ & $\textbf{80.65}{\pm.16}$\\
            $2$ & $73.54{\pm.28}$ & $70.60{\pm.41}$ & $\underline{73.63}{\pm.30}$ & $\textbf{74.06}{\pm.44}$ & $\underline{73.46}{\pm.43}$ & $71.91{\pm.22}$ & $73.26{\pm.18}$ & $\textbf{74.01}{\pm.33}$ & $73.37{\pm.17}$ & $71.51{\pm.50}$ & $\underline{73.75}{\pm.20}$ & $\textbf{73.87}{\pm.42}$\\
            $3$ & $\underline{77.57}{\pm.15}$ & $74.38{\pm.25}$ & $77.49{\pm.19}$ & $\textbf{77.93}{\pm.19}$ & $\underline{77.43}{\pm.14}$ & $74.88{\pm.22}$ &  $77.24{\pm.24}$ & $\textbf{77.80}{\pm.19}$ & $\underline{77.61}{\pm.09}$ & $76.20{\pm.33}$ & $77.30{\pm.10}$ & $\textbf{77.90}{\pm.25}$\\
            $4$ & $77.16{\pm.18}$ & $76.29{\pm.35}$ & $\underline{77.48}{\pm.17}$ & $\textbf{77.68}{\pm.15}$ & $77.21{\pm.20}$ & $76.28{\pm.45}$ & $\underline{77.54}{\pm.23}$ & $\textbf{77.94}{\pm.31}$ & $77.32{\pm.18}$ & $77.62{\pm.15}$ & $\underline{77.65}{\pm.25}$ & $\textbf{78.18}{\pm.22}$\\
            $5$ & $81.39{\pm.16}$ & $79.31{\pm.28}$ & $\textbf{81.79}{\pm.17}$ & $\underline{81.78}{\pm.25}$ & $81.46{\pm.15}$ & $80.68{\pm.44}$ & $\textbf{81.97}{\pm.13}$ & $\underline{81.75}{\pm.20}$ & $81.59{\pm.14}$ & $80.09{\pm.29}$ & $\textbf{81.78}{\pm.16}$ & $\underline{81.71}{\pm.23}$\\
            $6$ & $81.32{\pm.24}$ & $78.84{\pm.33}$ & $\textbf{81.48}{\pm.23}$ & $\underline{81.38}{\pm.28}$ & $81.23{\pm.34}$ & $79.84{\pm.38}$ & $\underline{81.30}{\pm.24}$ & $\textbf{81.60}{\pm.25}$ & $81.32{\pm.25}$ & $80.05{\pm.40}$ & $\underline{81.38}{\pm.17}$ & $\textbf{81.61}{\pm.25}$\\
            $7$ & $79.78{\pm.24}$ & $80.21{\pm.46}$ & $\underline{80.23}{\pm.19}$ & $\textbf{80.68}{\pm.43}$ & $80.00{\pm.11}$ & $\textbf{81.67}{\pm.20}$ & $80.32{\pm.20}$ & $\underline{80.46}{\pm.40}$ & $80.01{\pm.16}$ & $\textbf{82.18}{\pm.69}$ & $80.33{\pm.27}$ & $\underline{80.50}{\pm.34}$\\
            $8$ & $\underline{78.63}{\pm.12}$ & $75.15{\pm.46}$ & $78.60{\pm.20}$ & $\textbf{78.90}{\pm.29}$ & $78.82{\pm.30}$ & $74.66{\pm.49}$ & $\underline{78.94}{\pm.14}$ & $\textbf{79.07}{\pm.24}$ & $79.00{\pm.13}$ & $74.31{\pm.47}$ & $\underline{79.11}{\pm.11}$ & $\textbf{79.28}{\pm.30}$\\
            $9$ & $77.42{\pm.17}$ & $76.23{\pm.38}$ & $\underline{77.65}{\pm.21}$ & $\textbf{77.79}{\pm.35}$ & $77.56{\pm.18}$ & $77.26{\pm.55}$ & $\underline{77.65}{\pm.14}$ & $\textbf{77.96}{\pm.37}$ & $77.54{\pm.12}$ & $77.32{\pm.73}$ & $\underline{77.79}{\pm.15}$ & $\textbf{78.27}{\pm.30}$\\
            $10$ & $\underline{83.00}{\pm.21}$ & $80.93{\pm.30}$ & $\underline{83.00}{\pm.19}$ & $\textbf{83.19}{\pm.17}$ & $\underline{83.05}{\pm.16}$ & $82.47{\pm.34}$ & $82.89{\pm.17}$ & $\textbf{83.29}{\pm.21}$ & $\underline{83.21}{\pm.13}$ & $83.04{\pm.27}$ & $82.96{\pm.12}$ & $\textbf{83.31}{\pm.09}$\\
            $11$ & $78.79{\pm.22}$ & $\textbf{79.81}{\pm.36}$ & $79.05{\pm.22}$ & $\underline{79.22}{\pm.15}$ & $78.84{\pm.16}$ & $\textbf{81.13}{\pm.33}$ & $\underline{79.40}{\pm.15}$ & $79.30{\pm.22}$ & $79.18{\pm.08}$ & $\textbf{81.21}{\pm.49}$ & $79.40{\pm.11}$ & $\underline{79.54}{\pm.26}$\\
            $12$ & $\underline{74.13}{\pm.19}$ & $69.35{\pm.27}$ & $73.60{\pm.18}$ & $\textbf{74.34}{\pm.21}$ & $\underline{74.09}{\pm.20}$ & $71.36{\pm.33}$ & $73.82{\pm.37}$ & $\textbf{74.37}{\pm.32}$ & $\underline{74.16}{\pm.23}$ & $70.29{\pm.39}$ & $73.76{\pm.31}$ & $\textbf{74.49}{\pm.22}$\\
            $13$ & $78.03{\pm.17}$ & $\textbf{78.85}{\pm.43}$ & $78.18{\pm.24}$ & $\underline{78.33}{\pm.25}$ & $78.02{\pm.17}$ & $\textbf{78.66}{\pm.39}$ & $\underline{78.17}{\pm.15}$ & $77.79{\pm.23}$ & $78.09{\pm.11}$ & $\textbf{80.08}{\pm.59}$ & $\underline{78.23}{\pm.12}$ & $78.17{\pm.24}$\\
            $14$ & $80.79{\pm.19}$ & $77.53{\pm.46}$ & $\underline{80.85}{\pm.39}$ & $\textbf{80.98}{\pm.36}$ & $80.51{\pm.16}$ & $77.82{\pm.51}$ & $\underline{80.71}{\pm.22}$ & $\textbf{81.17}{\pm.29}$ & $80.44{\pm.27}$ & $77.52{\pm.32}$ & $\underline{80.66}{\pm.14}$ & $\textbf{80.89}{\pm.25}$\\
            $15$ & $\textbf{86.50}{\pm.19}$ & $83.87{\pm.40}$ & $86.20{\pm.13}$ & $\underline{86.31}{\pm.17}$ & $86.28{\pm.24}$ & $84.23{\pm.30}$ & $\underline{86.32}{\pm.20}$ & $\textbf{86.47}{\pm.23}$ & $\underline{86.43}{\pm.12}$ & $84.02{\pm.52}$ & $86.34{\pm.08}$ & $\textbf{86.80}{\pm.14}$\\
            $16$ & $79.48{\pm.25}$ & $75.13{\pm.47}$ & $\underline{79.82}{\pm.24}$ & $\textbf{80.03}{\pm.36}$ & $\underline{79.75}{\pm.29}$ & $74.81{\pm.31}$ & $\textbf{79.94}{\pm.35}$ & $\textbf{79.94}{\pm.19}$ & $79.82{\pm.12}$ & $75.24{\pm.82}$ & $\underline{80.08}{\pm.15}$ & $\textbf{80.11}{\pm.15}$\\
            $17$ & $76.84{\pm.25}$ & $70.56{\pm.36}$ & $\textbf{77.60}{\pm.22}$ & $\underline{77.30}{\pm.26}$ & $76.74{\pm.22}$ & $71.98{\pm.49}$ & $\underline{77.16}{\pm.20}$ & $\textbf{77.59}{\pm.34}$ & $76.84{\pm.13}$ & $72.08{\pm.46}$ & $\underline{77.23}{\pm.17}$ & $\textbf{77.47}{\pm.21}$\\
            $18$ & $\underline{60.64}{\pm.28}$ & $55.76{\pm.21}$ & $60.24{\pm.45}$ & $\textbf{61.09}{\pm.54}$ & $\underline{61.47}{\pm.41}$ & $58.74{\pm.23}$ & $61.46{\pm.34}$ & $\textbf{61.57}{\pm.35}$ & $\underline{61.87}{\pm.13}$ & $60.01{\pm.29}$ & $61.43{\pm.30}$ & $\textbf{62.28}{\pm.27}$\\
            $19$ & $74.19{\pm.22}$ & $73.77{\pm.77}$ & $\textbf{76.54}{\pm.47}$ & $\underline{75.51}{\pm.34}$ & $74.72{\pm.32}$ & $74.19{\pm.54}$ & $\underline{76.47}{\pm.31}$ & $\textbf{76.61}{\pm.57}$ & $74.66{\pm.30}$ & $74.42{\pm.54}$ & $\textbf{76.44}{\pm.24}$ & $\underline{76.17}{\pm.34}$\\
            $20$ & $61.63{\pm.32}$ & $56.81{\pm.27}$ & $\underline{62.34}{\pm.23}$ & $\textbf{62.45}{\pm.40}$ & $63.03{\pm.35}$ & $55.38{\pm.29}$ & $\underline{63.27}{\pm.32}$ & $\textbf{63.40}{\pm.33}$ & $\underline{63.21}{\pm.25}$ & $57.51{\pm.40}$ & $63.10{\pm.26}$ & $\textbf{63.23}{\pm.45}$\\
            $21$ & $\underline{82.65}{\pm.12}$ & $76.29{\pm.15}$ & $82.60{\pm.15}$ & $\textbf{82.67}{\pm.34}$ & $\underline{82.98}{\pm.10}$ & $76.30{\pm.15}$ & $82.73{\pm.16}$ & $\textbf{83.06}{\pm.20}$ & $\underline{83.04}{\pm.15}$ & $77.23{\pm.23}$ & $82.78{\pm.07}$ & $\textbf{83.06}{\pm.08}$\\
            \midrule
            AVG & $77.30{\pm.06}$ & $74.61{\pm.11}$ & $\underline{77.54}{\pm.04}$ & $\textbf{77.72}{\pm.05}$ & $77.46{\pm.06}$ & $75.34{\pm.07}$ & $\underline{77.64}{\pm.05}$ & $\textbf{77.89}{\pm.08}$ & $77.56{\pm.03}$ & $75.80{\pm.10}$ & $\underline{77.69}{\pm.05}$ & $\textbf{77.98}{\pm.07}$\\
			\bottomrule
	\end{tabular}}
\end{table*}
\begin{table*}
	\centering
	\caption{Complete AUC results on sentiment dataset for hyperparameters listed in Tab.~\ref{tab:param_mtgbdt}.}
	\label{tab:4tasks-10-25-50-100}
	\resizebox{0.985\textwidth}{!}{
		\begin{tabular}{c|cccc|cccc|cccc}
			\toprule
			\multirow{2}{*}{\textbf{Task}} & \multicolumn{4}{c|}{$\textbf{25\%}$} & \multicolumn{4}{c|}{$\textbf{50\%}$} & \multicolumn{4}{c}{$\textbf{100\%}$} \\
			& ~GBDT~~ & ~MT-ET~ & TSGB$_{\lambda}$ & TSGB & ~GBDT~~ & ~MT-ET~ & TSGB$_{\lambda}$ & TSGB & ~GBDT~~ & ~MT-ET~ & TSGB$_{\lambda}$ & TSGB\\
			\midrule
			Books & $88.56{\pm.14}$ & $\underline{88.89}{\pm.05}$ & $88.57{\pm.13}$ & $\textbf{89.47}{\pm.06}$ & $90.07{\pm.13}$ & $90.14{\pm.06}$ & $\underline{90.63}{\pm.14}$ & $\textbf{90.93}{\pm.06}$ & $93.86{\pm.13}$ & $92.79{\pm.06}$ & $\underline{93.97}{\pm.10}$ & $\textbf{94.37}{\pm.14}$\\
            DVDs & $88.89{\pm.11}$ & $88.78{\pm.04}$ & $\underline{89.21}{\pm.09}$ & $\textbf{89.56}{\pm.11}$ & $\underline{91.14}{\pm.08}$ & $90.57{\pm.05}$ & $91.02{\pm.09}$ & $\textbf{91.58}{\pm.08}$ & $94.14{\pm.09}$ & $92.26{\pm.05}$ & $\underline{94.24}{\pm.06}$ & $\textbf{94.39}{\pm.07}$\\
            Electr. & $93.47{\pm.10}$ & $\underline{93.97}{\pm.04}$ & $93.78{\pm.07}$ & $\textbf{94.45}{\pm.03}$ & $94.94{\pm.06}$ & $94.96{\pm.05}$ & $\underline{95.09}{\pm.09}$ & $\textbf{95.23}{\pm.04}$ & $95.98{\pm.08}$ & $95.66{\pm.03}$ & $\underline{96.03}{\pm.04}$ & $\textbf{96.06}{\pm.07}$\\
            K. App. & $94.43{\pm.07}$ & $94.47{\pm.03}$  & $\underline{94.65}{\pm.09}$ & $\textbf{95.09}{\pm.04}$ & $95.84{\pm.05}$ & $95.93{\pm.02}$ & $\underline{95.96}{\pm.09}$ & $\textbf{96.25}{\pm.04}$ & $96.99{\pm.03}$ & $96.52{\pm.03}$ & $\underline{97.11}{\pm.05}$ & $\textbf{97.20}{\pm.03}$\\
            \midrule
            AVG & $91.34{\pm.06}$ & $91.53{\pm.02}$ & $\underline{91.55}{\pm.04}$ & $\textbf{92.14}{\pm.04}$ & $93.00{\pm.04}$ & $92.90{\pm.02}$ & $\underline{93.18}{\pm.08}$ & $\textbf{93.50}{\pm.03}$ & $95.24{\pm.03}$ & $94.31{\pm.02}$ & $\underline{95.34}{\pm.04}$ & $\textbf{95.51}{\pm.05}$\\
			\bottomrule
	\end{tabular}}
\end{table*}

\end{document}